\documentclass[journal]{IEEEtran}
\ifCLASSINFOpdf
\else
\fi

\usepackage{graphicx}
\usepackage{rotating}
\usepackage{calc}
\usepackage[dvipsnames]{xcolor}
\usepackage[pagebackref=true,breaklinks=true,letterpaper=true,colorlinks,bookmarks=false]{hyperref}
\usepackage[hang,flushmargin]{footmisc}

\newcommand{\IText}[2]{
    \begin{sideways}
        \parbox{\heightof{#1}}{\centering #2} %
    \end{sideways}%
    \mbox{\hspace{.18\baselineskip} #1} %
}

\begin{document}
%
\title{A Unified Learning Based Framework for Light Field Reconstruction from Coded Projections}

\author{Anil~Kumar~Vadathya \thanks{\noindent A. Vadathya is with the Department of Electrical and Computer Engineering, Rice University, Houston, TX 77005},~
        Sharath~Girish \thanks{\noindent S. Girish is with the Department of Computer Science, University of \newline Maryland, College Park, MD 20740},~
        and~Kaushik~Mitra,~\IEEEmembership{Member,~IEEE} \thanks{\noindent K. Mitra is with the Department of Electrical Engineering, Indian Institute of Technology Madras, Chennai 600036}}
\maketitle

\begin{abstract}
Light fields present a rich way to represent the 3D world by capturing the spatio-angular dimensions of the visual signal. However, the popular way of capturing light fields (LF) via a plenoptic camera presents a spatio-angular resolution trade-off. Computational imaging techniques such as compressive light field and programmable coded aperture  reconstruct full sensor resolution LF from coded projections obtained by multiplexing the incoming spatio-angular light field. Here, we present a unified learning framework that can reconstruct LF from a variety of multiplexing schemes with \textit{minimal} number of coded images as input. We consider three light field capture schemes: heterodyne capture scheme with code placed near the sensor, coded aperture scheme with code at the camera aperture and finally the dual exposure scheme of capturing a focus-defocus pair where there is no explicit coding. Our algorithm consists of three stages 1) we recover the all-in-focus image from the coded image 2) we estimate the disparity maps for all the LF views from the coded image and the all-in-focus image, 3) we then render the LF by warping the all-in-focus image using disparity maps.
We show that our proposed learning algorithm performs either on par with or better than the state-of-the-art methods for all the three multiplexing schemes. 
LF from focus-defocus pair is especially attractive as it requires no hardware modification and produces LF reconstructions that are comparable to the current state of the art learning-based view synthesis approaches from multiple images. 
Thus, our work paves the way for capturing full-resolution LF using conventional cameras such as DSLRs and smartphones.
\end{abstract}

\begin{IEEEkeywords}
Light field resolution trade-off, compressive light field imaging, coded aperture photography, disparity based view synthesis.
\end{IEEEkeywords}

%
\IEEEpeerreviewmaketitle

{\let\thefootnote\relax\footnote{{This paper has supplementary downloadable material available at http://ieeexplore.ieee.org., provided by the author. The material includes the animations of reconstructed light fields. This material is 93MB in size.}}}

\section{Introduction}
\IEEEPARstart{L}{ight} field imaging has renewed interests for 3D modeling with the availability of commercial light field cameras, such as, Lytro and Raytrix.
By capturing angular rays arriving from a scene point it offers post capture control over focal plane and different perspectives of the scene. However, capturing the 4D light field on a 2D sensor presents \textit{resolution trade-off}, where, usually the spatial resolution is sacrificed in order to accommodate the angular resolution. 

To address this LF resolution trade-off issue, a wide variety of solutions have been proposed such as super-resolution techniques \cite{schedl2015directional, mitra2012light, shi2014light}, view synthesis approaches \cite{wanner2014variational, georgiev2006spatio, kalantari2016learning} and computational cameras with specialized optical multiplexing hardware \cite{ashok2010compressive, marwah2013compressive, liang2008programmable, babacan2012compressive, inagaki2018learning}. View synthesis approaches enhance angular resolution from a sparse set of views by synthesizing the novel views via disparity estimation. Computational imaging approaches optically multiplex the incoming spatio-angular light rays to 2D coded projections on the sensor and reconstruct light field back from these projections. 
Other methods work by reconstructing 4D LF from 3D focal stack \cite{levin2010linear, mousnier2015partial}. 

More recently, data driven techniques using deep networks have been proposed for addressing resolution trade-off. For example, \cite{kalantari2016learning} use four corner views of a $8\times8$ light field to interpolate the intermediate views by employing a learning based view synthesis approach. However, note that this approach needs access to a sparse set of views e.g. four corner views from the light field for enhancing its angular resolution, thereby making it difficult to integrate it with regular consumer cameras. 
Another learning based technique \cite{srinivasan2017learning} attempts to reconstruct LF from a single image, which is an  extremely ill-posed problem. They show visually convincing results for a particular dataset of flowers. However, since single image lacks geometric cues, the approach doesn't generalize well for other scenes. 

Inspired by the success of deep networks for light field reconstruction,
we leverage deep networks for reconstructing light field from coded projections. For this we consider three frameworks that broadly span the spatio-angular multiplexing schemes. The first scheme is \textit{compressive light field} (CLF) imaging, where a code is placed near to the sensor \cite{marwah2013compressive}. The modulation along both the spatial and angular dimensions enables LF reconstruction from a single coded image. The second scheme that we consider is the \textit{coded aperture} (CA) imaging with the code placed at the camera aperture \cite{levin2007image, veeraraghavan2007dappled}. With modulation along angular dimension alone, this scheme usually requires multiple coded images as input \cite{liang2008programmable, babacan2012compressive, inagaki2018learning}. However, we show that using our deep learning approach, we are able to reconstruct LF from a single coded image.

Relaxing the hardware requirements further, for the first time, we propose light field reconstruction from two images captured using a conventional camera by varying the camera aperture. This brings us to our third scheme of LF reconstruction from a pair of \textit{all-in-focus} and \textit{defocus} image, where we exploit the relative defocus cue between the two images. The main advantage of focus-defocus scheme over those of CA and CLF is that we can capture these images using just a conventional camera by varying its aperture. Note that reconstructing LF from just a single defocused image is extremely difficult. The angular modulation function here is a simple averaging operation and hence it is not as well-conditioned as others for inversion. 

\subsection{Unified framework for light field reconstruction}

\begin{figure*}[t]
    \centering
    \includegraphics[width=0.9\textwidth]{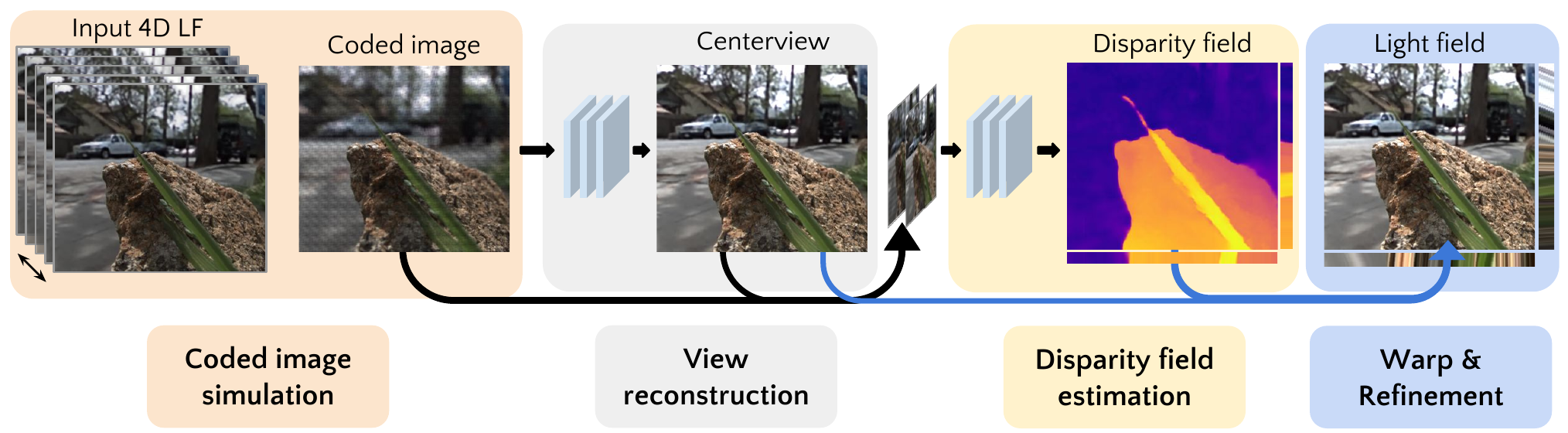}
    \caption{Our disparity based light field reconstruction framework from coded projections. \textit{Coded image simulation:} We simulate the coded image from the input 4D LF as discussed in Sec. \ref{sec:acquiring_ic}. \textit{View reconstruction:} ViewNet reconstructs centerview from the input coded image. \textit{Disparity field estimation:} DisparityNet estimates disparity field (disparity at every view-point in the light field) from the input concatenated coded image and the centerview. \textit{Warp \& Refinement:} Now we warp the centerview using disparity field to generate warped LF which is further refined to account for occlusion information.}
    \label{fig:pipeline}
\end{figure*}

We take advantage of the fact that in all the multiplexing schemes the geometric or depth cue is explicitly encoded in the coded projections. Thus, we leverage on this cue to propose a disparity based light field reconstruction algorithm, where we 
first estimate the depth or disparity map and then use this to reconstruct the LF. 
The disparity estimation step and the LF reconstruction (or warping) step are common for all the three coding schemes and hence we propose an unified framework for these steps, (see Fig. \ref{fig:pipeline}). 


More specifically, our unified LF reconstruction framework consists of the following steps. The first step is \textit{view reconstruction}, where we reconstruct the centerview of the light field from the coded projection. The second step is \textit{disparity estimation}, where by using the relative information between the reconstructed centerview and the coded projection, we estimate the \textit{disparity field}. The disparity field is basically disparity map at all views of the light field. Disparity field enables us to perform seamless backward warping of the centerview to all the novel view points in the light field. Once we have both the centerview and disparity field, we warp the center view using the disparity field to generate the Lambertian light field. Since warping the centerview does not account for recovering the occluded pixels, we further refine the light field using a residual learning block along the lines of \cite{srinivasan2017learning}. For the tasks of view reconstruction and disparity estimation, we use two deep neural networks - \textit{ViewNet} and \textit{DisparityNet} in our pipeline (see Fig. \ref{fig:pipeline}). ViewNet is adapted from an earlier work of image restoration by Mao et al. \cite{mao2016image}. DisparityNet is inspired from the disparity estimation architecture of Srinivasan et al. \cite{srinivasan2018aperture} to which we add an additional block for multiscale feature aggregation. This helps our disparity estimation in ambiguous regions and improves our LF reconstruction performance by 1dB as we show later in ablation study.

\begin{figure*}[t]
    \centering
    \begin{minipage}{0.7\textwidth}
        \centering
        \includegraphics[width=\textwidth]{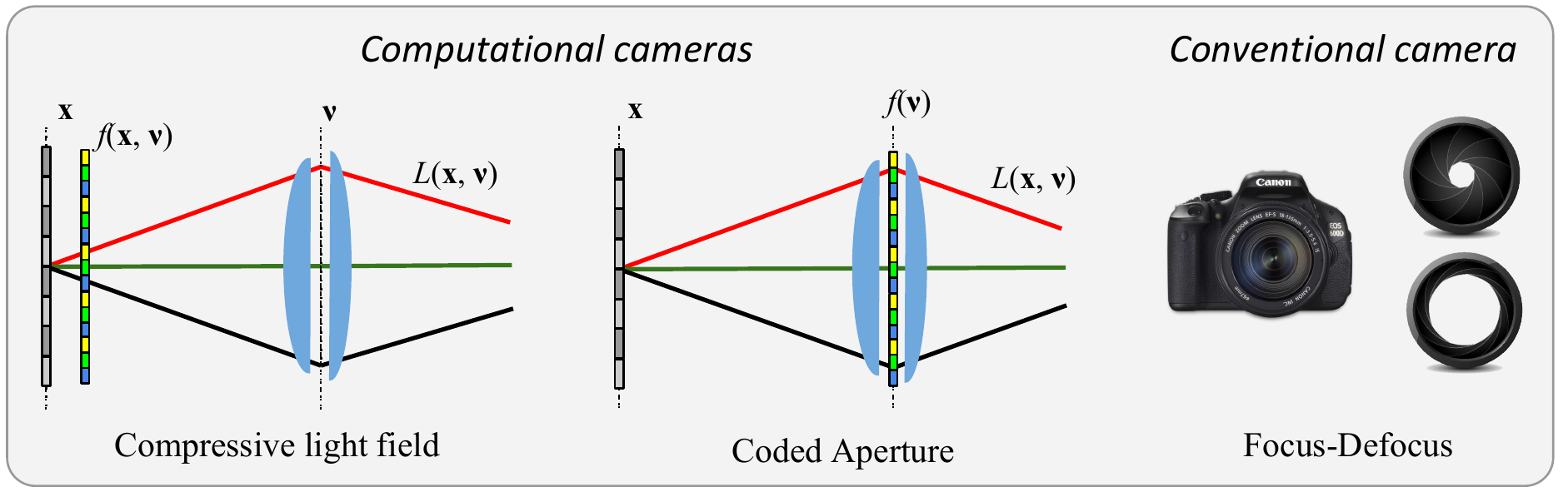}
    \end{minipage}
    \begin{minipage}{0.22\textwidth}
        \centering Seahorse scene  
        \includegraphics[width=\textwidth]{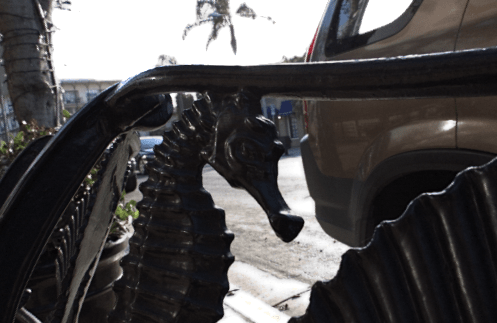}
    \end{minipage}
    \\\vspace{0.25cm}
    \begin{minipage}{0.155\textwidth}    \centering  \small  Single image \cite{srinivasan2017learning} \end{minipage}
    \begin{minipage}{0.155\textwidth}    \centering  \small  View synthesis \cite{kalantari2016learning} \end{minipage}
    \begin{minipage}{0.155\textwidth}    \centering  \small  Ours \\ Compressive LF \end{minipage}
    \begin{minipage}{0.155\textwidth}    \centering  \small  Ours \\ Coded  Aperture \end{minipage}
    \begin{minipage}{0.155\textwidth}    \centering  \small  Ours \\ Focus-Defocus     \end{minipage}
    \begin{minipage}{0.155\textwidth}    \centering  \small  From full 4D LF \cite{shin2018epinet} \end{minipage}
    \\\vspace{0.1cm}
    \includegraphics[width=.157\textwidth]{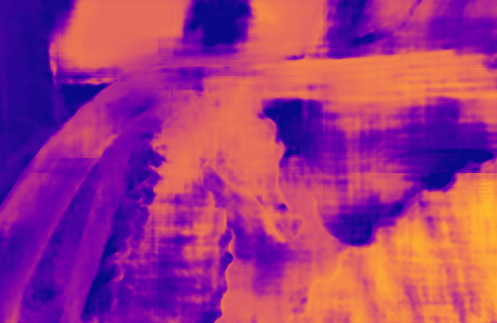}
    \includegraphics[width=.157\textwidth]{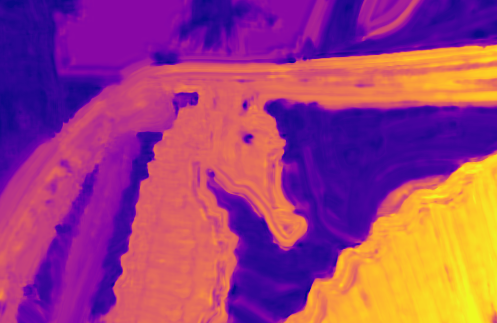}
    \includegraphics[width=.157\textwidth]{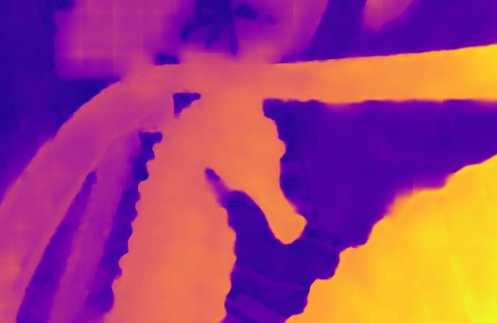}
    \includegraphics[width=.157\textwidth]{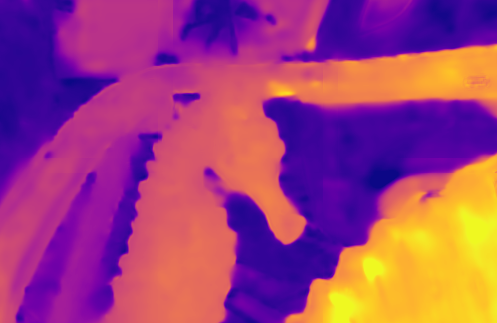}
    \includegraphics[width=.157\textwidth]{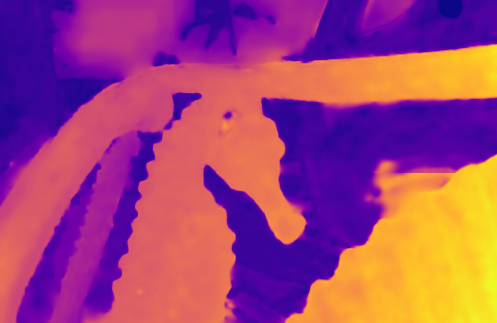}
    \includegraphics[width=.157\textwidth]{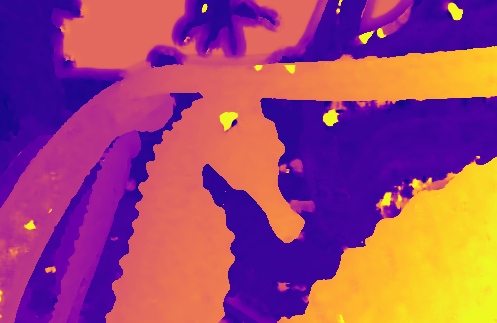}
    \caption{
    We propose a unified learning based framework for full sensor resolution light field (LF) reconstruction from coded projections with as minimal input as possible. We consider three frameworks that broadly span the spatio-angular multiplexing spectrum: Compressive light field (CLF), coded aperture (CA) and focus-defocus pair (FocDef). We propose to do disparity based LF reconstruction where disparity estimation is learned in an unsupervised manner. Our analysis shows that focus-defocus with \textit{minimal} hardware requirements performs as good as the hardware intensive approaches of CLF and CA. Focus-defocus is also slightly better than the four corner views approach of Kalantari et al. \cite{kalantari2016learning}. Thus, our reconstruction framework from focus-defocus pair paves the way for capturing high resolution LFs using a conventional camera. }
    \label{fig:teaser}
\end{figure*}

Using the proposed unified framework, we compare our LF reconstruction with state-of-the-art methods in each of the multiplexing schemes. For compressive light field reconstruction (CLF) we compare our results with dictionary learning approach of Marwah et al. \cite{marwah2013compressive} and direct regression approach of Nabati et al. \cite{nabati2018fast}. For the case of coded aperture (CA) we compare our results with the recent learning based approach of Inagaki et al. \cite{inagaki2018learning} which directly reconstructs LF from coded image without any geometry estimation. Since there is no previous work in reconstructing LF from focus-defocus (FocDef) pairs we compare this with recent learning based view synthesis approaches such as the four corner views method of Kalantari et al. \cite{kalantari2016learning} and the single image based LF reconstruction approach of  Srinivasan et al. \cite{srinivasan2017learning}. Our experiments show that in all these three multiplexing schemes, our disparity based LF reconstruction pipeline performs either on par with or better than the state-of-the-art. For CLF, our disparity based LF reconstruction outperforms the traditional dictionary learning approach \cite{marwah2013compressive} and learning approach \cite{nabati2018fast} which directly reconstructs LF without any intermediate disparity estimation. In case of CA, our method using just one coded input produces LF reconstructions that are on, average, 2dB better than the competing method of \cite{inagaki2018learning}. Our LF from focus-defocus approach is especially attractive as with no hardware modification, it produces better LF reconstructions than CLF and CA. Foc-Def also produces slightly better results than the state-of-the-art  learning-based view synthesis approaches from multiple images. Finally, we demonstrate that focus-defocus enables us to reconstruct full sensor resolution LF from just two images captured using a conventional DSLR camera (see Fig. \ref{fig:real_lf}).

Our contributions are as follows:
\begin{itemize}
    \item We propose a unified deep learning based framework for LF reconstruction from coded measurements. Our main step is the disparity estimation network \textit{DisparityNet}, which we train using indirect supervision from view-synthesis. 
    \item Using our reconstruction algorithm we analyze three different LF reconstruction frameworks 1) compressive light field (CLF) 2) coded aperture (CA) and 3) focus-defocus (FocDef) image pair. \textcolor{black}{We believe that this is the first work to look at LF reconstruction from focus-defocus image pair.}
    \item With our unified LF reconstruction pipeline we show that we are on par with the state-of-the-art techniques in all the multiplexing schemes. Our FocDef approach with no additional hardware requirement performs better than hardware intensive CLF and CA. Thus, it paves the way for capturing LFs with conventional cameras.
    \item Finally, we demonstrate that we can reconstruct high resolution light field from real focus-defocus image pair captured using a DSLR. 
\end{itemize}

\section{Related work}\label{sec:related_work}

\textit{Light field imaging}. Light field data \cite{lippmann} captures the spatio-angular information, that is, light rays coming at different angles from a point in the scene.  By capturing the spatio-angular information light field enables synthesis of novel views \cite{levoy1996light}, post-capture control over focal plane and aperture \cite{ng2005light}. Early works used camera arrays \cite{levoy1996light} to capture light fields which are bulky for regular use. Recently, lenset based approaches \cite{ng2005light} for capturing LFs have been successfully adopted into consumer cameras like Lytro \cite{lytro} and Raytrix \cite{raytrix}. Due to limited sensing resolution, capturing light fields present a \textit{trade-off} between the spatial and angular resolution. 

\textit{Super-resolution of light fields}. \textcolor{black}{Many initial approaches propose to do angular super-resolution of light fields using a suitable prior \cite{schedl2015directional, mitra2012light, shi2014light}. 
Zhang et al. \cite{zhang2015light} and Didyk et al. \cite{didyk2013joint} propose phase based approaches to synthesize light fields from the input stereo pairs. Recently, neural network based approaches, such as Yoon et al. \cite{yoon2015learning}, Gu et al. \cite{gul2018spatial}, Wu et al. \cite{ wu2017light} train deep convolutional networks (CNN) for spatio-angular super-resolution.} 

\textit{Light field from coded projections}. Many computational imaging techniques were proposed for light field reconstruction from coded projections of spatio-angular rays. Liang et al. \cite{liang2008programmable} propose LF reconstruction from a programmable aperture camera with optimal multiplexing patterns. The number of coded images they require as input is as high as the LF angular resolution itself. Ashok et al. \cite{ashok2010compressive} suggest that by exploiting the spatio-angular correlations inherent in LFs, they can be reconstructed from much less measurements than implied by their angular resolution. 
Babacan et al. \cite{babacan2012compressive} use programmable aperture to do LF reconstruction from coded images with reduced number of measurements than \cite{liang2008programmable}. 
Marwah et al. \cite{marwah2013compressive} realize the hardware for compressive light field camera by placing a mask near the sensor. This spatio-angular multiplexing enables them to reconstruct \textit{full sensor resolution} LF from just a single coded image via LF dictionary and optimal projections. \textcolor{black}{Hirsch et al. \cite{hirsch2014switchable} propose a switchable camera with angle sensitive pixels which can capture high resolution 2D images and 4D light fields}. In addition, other set of works explore reconstructing 4D light field from 3D focal stack. For this purpose, Levin et al. \cite{levin2010linear} use dimensionality gap prior and Mousnier et al. \cite{mousnier2015partial} use tomography based back-projection. Instead of reconstructing light field from focal stack which uses tens of images, in our third scheme, we propose to reconstruct LF from just a pair of focus-defocus images. 
\textit{Disparity based view synthesis for light field cameras}. These approaches involve two steps for view synthesis. First step involves estimating disparity at the novel view point. Second step involves warping the available views and weighing them appropriately to finally render the novel view. Wanner et al. \cite{wanner2014variational} propose a variational framework for disparity estimation for novel view synthesis to enhance LF angular resolution. Flynn et al. \cite{flynn2016deepstereo} use deep neural networks for novel view synthesis from a set of camera views with wide baseline. 
Kalantari et al. \cite{kalantari2016learning} propose a learning based view synthesis for light field cameras. They propose to do full light field ($8\times8$) reconstruction from just four corner sub-aperture images. Inspired from the recent works of unsupervised geometry estimation \cite{godard2017unsupervised, garg2016unsupervised, flynn2016deepstereo}, they propose to learn disparity using the view-synthesis loss. By simultaneously learning both the disparity estimation and color prediction they show visually convincing results. This is specially so around occlusion edges where earlier approaches of Wanner et al. \cite{wanner2014variational} exhibit tearing artifacts. 
Our approach differs from these view synthesis approaches basically in the form of input for LF reconstruction. These approaches typically have multiple views as input whereas our framework, in case of CLF and CA, uses just a single coded image.  


\textit{Single image based LF synthesis}. Srinivasan et al. \cite{srinivasan2017learning} attempt to reconstruct a 4D light field from a single RGB image. They propose a disparity based approach where disparity is estimated from a single image using a deep network. Using this disparity map, they warp the single image to obtain a Lambertian LF. This is followed by a
refinement step to account for non-Lambertian effects like occlusions. Since single image geometry estimation is ill-posed, they train their method on specific category of scenes. In contrast, our framework exploits explicit geometric cues to synthesize the light field from the coded image and hence can generalize to a variety of scenes.

\textit{Depth from coded projections}. Many works were proposed for estimating depth and all-in-focus image from coded projections such as  \cite{levin2007image, veeraraghavan2007dappled, masia2012perceptually}. Depth from focus-defocus pairs is also well studied in the depth from defocus (DfD) literature  \cite{pentland1987new, rajagopalan1999mrf, hasinoff2009multiple}. In our coded aperture case similar to Levin et al. \cite{levin2007image} we also recover image and depth from a single coded capture but with the goal of reconstructing light field. In focus-defocus case, disparity estimation requires disambiguation of the depth planes in-front of focal plane with the depth planes behind it. DfD works do not account for this and typically require the focal plane to be on the nearest depth plane. The multi-aperture imaging work of Hasinoff et al. \cite{hasinoff2009multiple} proposes a greedy approach for depth disambiguation for restricted set of scenes. Our disparity estimation network on the other hand learns to do the depth disambiguation via light field reconstruction loss. 

In this work, we extend upon the work of \cite{vadathya2018learning} which does disparity based compressive LF reconstruction. We extend it to another popular multiplexing scheme, which is LF reconstruction from a single coded aperture image, and also propose a third scheme, which is LF reconstruction from focus-defocus image pair. From the algorithm point of view, \cite{vadathya2018learning} predict only the centerview disparity map in the light field. Thus, for rendering light field they use forward warping which requires hole-filling in the novel view. On the other hand, we estimate the disparity field, disparity map at every view of LF, which enables seamless backward warping with no requirement for hole-filling. 
Also, we refine the warped light field to account for occlusion. Further we analyze the performance of the three schemes, compare them with state-of-art LF reconstruction algorithms and finally demonstrate our algorithm on real focus-defocus pairs captured from a DSLR.

\section{Acquiring Coded projections} \label{sec:acquiring_ic}
As we discussed, we consider three frameworks for light field reconstruction from coded projections. Here we describe mathematically the projection of incident spatio-angular light field to a 2D coded image in detail for all these setups. For this, we follow the two plane parametrization for spatio-angular light field $L(\mathbf{x}, \mathbf{v})$, where $\mathbf{x}$ and $\mathbf{v}$ are the 2D spatial co-ordinates on the sensor plane and 2D angular position on the aperture plane, respectively.

\subsection{Compressive light field (CLF) photography}
Here, we consider the optical multiplexing proposed by \cite{marwah2013compressive}. This involves placing the coded mask near the sensor and in between the sensor and lens (see Fig. \ref{fig:teaser}). The coded image formation is given as follows, 
\begin{equation}
    \label{eq:clf_ic}
    I_c (\mathbf{x}) = \int_ {\mathbf{v}} f(\mathbf{x}, \mathbf{v}) L(\mathbf{x},  \mathbf{v}) d\mathbf{v},
\end{equation}
where the incoming spatio-angular light field is modulated by the mask $f$ along both spatial and angular dimensions. Due to this, each position on the sensor plane observes a different projection of angular rays. Thus, the spatio-angular modulation provides best case of acquiring incoherent measurements. This enables for a 4D light field reconstruction from a single 2D coded projection \cite{xu2012high, marwah2013compressive}. 

\subsection{Coded aperture (CA) light field photography}
Coded aperture is a further simplification of multiplexing with coded mask being moved on to the aperture \cite{levin2007image, veeraraghavan2007dappled}. The coded image formation is as follows: 
\begin{equation}
    \label{eq:ca_ic}
    I_c (\mathbf{x}) = \int_\mathbf{v} f(\mathbf{v}) L(\mathbf{x}, \mathbf{v})d\mathbf{v},
\end{equation}
with modulation only along the angular dimension. Since the coded mask is on aperture plane now, essentially, each pixel on the sensor plane observes the same projection of angular rays. This provides certain amount of hardware relaxation than compressive LF at the cost of multiplexing being spatially invariant. Thus, coded aperture frameworks \cite{babacan2012compressive} accommodate for this with multiple measurements for light field reconstruction.  

\subsection{Focus-defocus (FocDef) pair}
Here we relax the explicit coded projections in the above cases and propose to use the relative defocus cue between two images introduced by aperture variation for light field reconstruction. The defocus image can be thought of as a special case of coded aperture imaging with code $f$ being uniform and the in-focus image as the pin-hole view. The aperture variation cue relaxes to a greater degree the specialized hardware necessary for light field reconstruction. Thus, paving the way for high resolution light field reconstruction with conventional cameras like DSLR. 

\section{Light field from coded projections}
Our goal is to reconstruct a 4D light field $L(\mathbf{x}, \mathbf{v})$ from its 2D coded projection $I_c(\mathbf{x})$. We propose a deep learning architecture to learn this mapping from the coded image to the light field. 
Instead of directly reconstructing light field from coded projections, we break it down into \textit{view reconstruction} and \textit{disparity estimation} steps. By enforcing disparity estimation as an intermediate step for LF reconstruction our approach provides an interpretable solution. The estimated disparity values directly translate to slopes of epi-polar image (EPI) lines in the reconstructed LF. Whereas, in case of direct reconstruction no such interpretation is possible. The neural network has to explicitly learn to extract the geometry  which is not very efficient as shown by recent works of \cite{flynn2016deepstereo, kalantari2016learning, srinivasan2017learning}. 

We breakdown the light field reconstruction from coded images into three stages. In the first stage we train a deep network for reconstructing the centerview of the light field $L(\mathbf{x}, 0)$ from the coded image $I_c(\mathbf{x})$: 
\begin{equation}
    \hat{L}(\mathbf{x}, 0) = v(I_c(\mathbf{x})), 
\end{equation}
where $v$ is a deep neural network which we call as \textit{ViewNet}, and $\hat{L}(\mathbf{x}, 0)$ is the predicted centerview. In the second stage, we train a deep network to estimate the disparity field $D(\mathbf{x}, \mathbf{v})$, that is, disparity map for every spatial position and angular view $(\mathbf{x}, \mathbf{v})$ in the light field. Note that if we predict the centerview disparity alone instead of the whole disparity field then we would have to employ forward warping of centerview which might leave holes in the interpolated image. We leverage the relative information between the coded image $I_c(\mathbf{x})$ and reconstructed centerview $\hat{L}(\mathbf{x}, 0)$ to estimate the disparity field:
\begin{equation}
    D(\mathbf{x}, \mathbf{v}) = d\big(\hat{L}(\mathbf{x}, 0), I_c(\mathbf{x})\big),
\end{equation}
where $d$ is a deep network which we call
as \textit{DisparityNet}. The disparity map at a view point is defined with respect to its immediate neighbour. Note that estimating disparity maps at novel views would require the estimation of disparity values for missing pixels around occlusion edges which are not present in the centerview. Finally, in the third stage, we perform light field synthesis by backward warping the centerview $\hat{L}(\mathbf{x}, 0)$ according to the disparity map $D(\mathbf{x}, q)$ at viewpoint $q$ giving us the warped light field $\hat{L}(\mathbf{x}, q)$ at viewpoint $q$. We repeat this process for all viewpoints to obtain the complete LF.
\begin{equation}
    \hat{L}(\mathbf{x}, q) = \hat{L}(\mathbf{x} + qD(\mathbf{x}, q), 0).
\end{equation}
Note, since the disparity map is with respect to its immediate neighbor we need to multiply it with $q$ to get disparity with respect to the centerview. We train the disparity based light field rendering in an end to end manner by minimizing the light field reconstruction error. We use bilinear interpolation for warping and hence the entire pipeline is differentiable. Thus, \textit{DisparityNet} learns the scene geometry from light field reconstruction loss without the explicit need for ground-truth disparities. The LF reconstruction loss is given by:  
\begin{equation}
    \mathcal{L}_{rec} =\| \hat{L}(\mathbf{x}, \mathbf{v}) - L(\mathbf{x}, \mathbf{v}) \|_1.
    \label{eq:loss_rec}
\end{equation}

\subsection{Disparity from coded projections}

In order to estimate disparity from the coded images, we propose to learn the disparity using view-synthesis loss (Eqn. \ref{eq:loss_rec}). Although, an alternative approach would be to compute the depth map first using conventional approaches, for example, for focus-defocus pair we can use depth from defocus approaches and then synthesize the LF from that. However, this approach of separating the disparity estimation and synthesis of novel views is shown to be ineffective in earlier works of Kalantari et al. \cite{kalantari2016learning}, Wu et al. \cite{wu2017light} as it leads to tearing artifacts around occlusion edges. Thus, an end-to-end training for disparity estimation and view synthesis is preferable. However, learning to estimate disparity poses certain challenges which are discussed below.

\subsubsection{Challenges with disparity estimation}
\label{ss_sec:disp_estim_challenges} 
First of all, for constant intensity regions, any disparity value is likely to reduce the reconstruction error. Hence, there is an ambiguity as to the correct disparity value for such regions. 
Second challenge is in estimating the correct disparity sign for scene depth planes in-front (+ve disparity) or behind (-ve disparity) the focal plane. For example, in case of focus-defocus pair, both the depth planes in front and behind the focal plane see similar defocus blur. Only the depth edges contain the cue as to the ordering of depth planes. 






\begin{figure*}[t]
    \centering
    \includegraphics[width=0.95\textwidth]{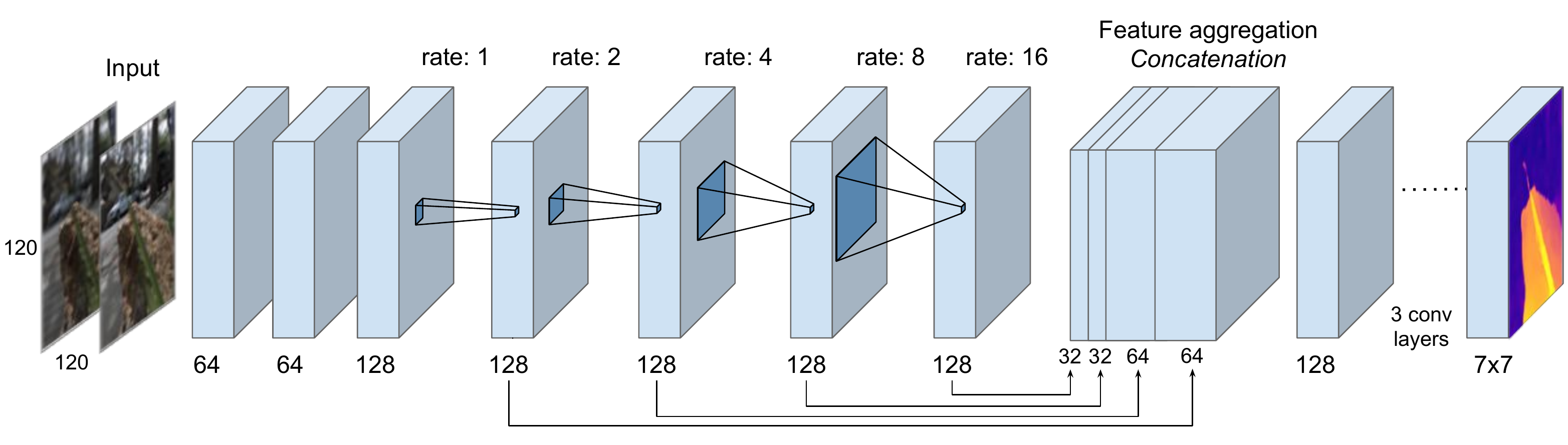}
    \caption{\textit{DisparityNet}: Network for disparity estimation from coded projections. We use a $3\times3$ conv. filter at all layers followed by ELU activation function and \textit{batch normalization} except for the last layer which outputs view disparities ($7\times7$). We employ dilated convolutions for expanding the receptive fields with the dilation rate specified at each layer. Our feature aggregation step concatenates multi-scale features obtained with different dilation rates, as indicated by the arrows below. Notice that we reduce the number of channels of features while concatenating them for the feature aggregation. Each of the arrows indicated performs a two layer convolution to reduce the number of channels. For example, features of 128 channels of rate$=16$ are reduced to 32 channels for aggregation. Feature aggregation step is followed by five more layers of convolutions, finally, predicting view disparities. For further details, see Sec. \ref{ssec:dispnet}.}
    \label{fig:dispnet}
\end{figure*}

To address these problems, we introduce a multiscale feature aggregation block in our disparity field estimation network shown in Fig. \ref{fig:dispnet}. We use dilation rates in our convolutions to expand the receptive field, that is, the region in the input image that affects the filter output. Thus, dilated convolutions allow us to leverage neighbourhood context while estimating disparity values. We additionally concatenate the features from different dilation rates allowing us to leverage context around a pixel at multiple scales. This feature aggregation step also helps us in estimating the relative position with respect to the focal plane. As we will see in our results later, disparity estimation without such feature aggregation results in erroneous predictions (see Fig. \ref{fig:noMS}). 

In addition, for better disambiguation between positive and negative disparities with respect to the focal plane, we additionally augment our training data by shearing the light field to vary the plane of focus in the coded images. This ensures that the disparity estimation network sees a variety of relative focal plane positions while learning the parameters. 

\subsubsection{Disparity consistency regularization}
\label{sssec:dc}
As discussed earlier, our \textit{DisparityNet} predicts disparity field which is disparity map at ever view point or sub-aperture image of the light field. Note that for the disparity maps across the views to be consistent, all the views should have same disparity value for a 3D point in the scene. To enforce such consistency among disparity maps in a disparity field $D(\mathbf{x}, \mathbf{v})$, we additionally enforce photo-consistency among disparity maps. For example, disparity at views $\mathbf{v}$ and $\mathbf{v}+q$ should satisfy, $D(\mathbf{x},\mathbf{v}) = D(\mathbf{x} - qD(\mathbf{x}, \mathbf{v}), \mathbf{v}+q)$ which leads to the following regularization, 

\begin{equation}
    \mathcal{L}_{dc} = \| D(\mathbf{x},\mathbf{v}) - D(\mathbf{x} - qD(\mathbf{x}, \mathbf{v}), \mathbf{v}+q) \|_1.
    \label{eq:loss_epi}
\end{equation}

Godard et al. \cite{godard2017unsupervised} and Srinivasan et al. \cite{srinivasan2017learning} show that using such geometric consistencies among the disparity maps improves the disparity estimation over using the view reconstruction loss alone.

\textbf{Loss function} Overall our loss function is given by, 
\begin{equation}
    \sum_{i, d_i} \mathcal{L}_{rec}(L_i, \hat{L}_i) +  \lambda_{dc}\mathcal{L}_{dc}(D_i) + \lambda_{tv}\mathcal{L}_{tv}(D_i),
    \label{eq:overall_loss}
\end{equation}
where, $\mathcal{L}_{dc}$ ensures sparsity of $D(\mathbf{x}, \mathbf{v})$ along EPI lines, $\mathcal{L}_{tv}$ \textcolor{black}{ensures spatial sparsity of the disparity map for each view}. The term $d_i$ indicated in the sum refers to shearing of LF in each batch to change the focal plane in the scene resulting in variety of relative depths (see Sec. \ref{ss_sec:disp_estim_challenges}). In addition to this loss function we also have a refinement loss proposed by Srinivasan et al. \cite{srinivasan2018aperture}. The refinement stage takes as input the warped LF, disparity field and predicts a residual LF which is added to the warped LF resulting in the refined LF which is then compared to the ground truth.

In the following section, we will discuss the individual neural networks in detail.

\section{A Unified Learning Framework}
\subsection{ViewNet: View reconstruction} \label{ssec:view_reconst}
As discussed earlier, the first step in our light field reconstruction framework is recovering the centerview of the light field from the coded projection. 
View reconstruction involves recovering high frequecy details from the coded image. We pose this as an image restoration problem and use the state-of-the-art approach of \cite{mao2016image}. Note that view reconstruction is necessary only for the cases of CLF and CA. In the case of focus-defocus pair,  the all-in-focus input serves as centerview. 

\textit{ViewNet} consists of encoder-decoder architecture with symmetric skip connection to preserve the low-level details. We use 15 such symmetric skip connections for centerview reconstruction for both the cases of CLF and CA. We use $3\times3$ filters in the CNN. The network is trained by minimizing the $l_1$ reconstruction error. 

\subsection{DisparityNet: Disparity field estimation}
\label{ssec:dispnet}
Our next step involves estimating the scene geometry from the reconstructed centerview and the coded image for rendering the light field. 
We concatenate both the coded image and centerview as input to our disparity estimation network for predicting the disparity field, $D(\mathbf{x}, \mathbf{v})$. Although the disparity could be estimated from the coded projection alone, using relative information between the coded image and centerview makes it easier. Moreover, the centerview is anyway available from the view reconstruction step as part of LF reconstruction. 

\textit{DisparityNet} shown in Fig. \ref{fig:dispnet} estimates disparity at all the views in the light field. The network uses dilated convolutions \cite{yu2015multi} with successive dilation rates of 2, 4, 8 and 16 
which gives a receptive field of size $67\times67$ in the input image. Then, we concatenate features from different dilation rates which combines information at different scales and is followed by few more convolutional layers to output disparity maps at all the views, in this case $7\times7$. Each convolution layer is followed by ELU activation and batch normalization. The last layer is followed by a \textit{Tanh} activation layer with a scaling factor of 10, to limit the disparities between adjacent views to $[-10,10]$. The output disparity maps are used to generate the warped light field. In addition to the reconstruction loss, disparity maps are also regularized for disparity consistency as discussed in Sec. \ref{sssec:dc}.

\begin{figure*}[t]
    \centering
    \includegraphics[width=\textwidth]{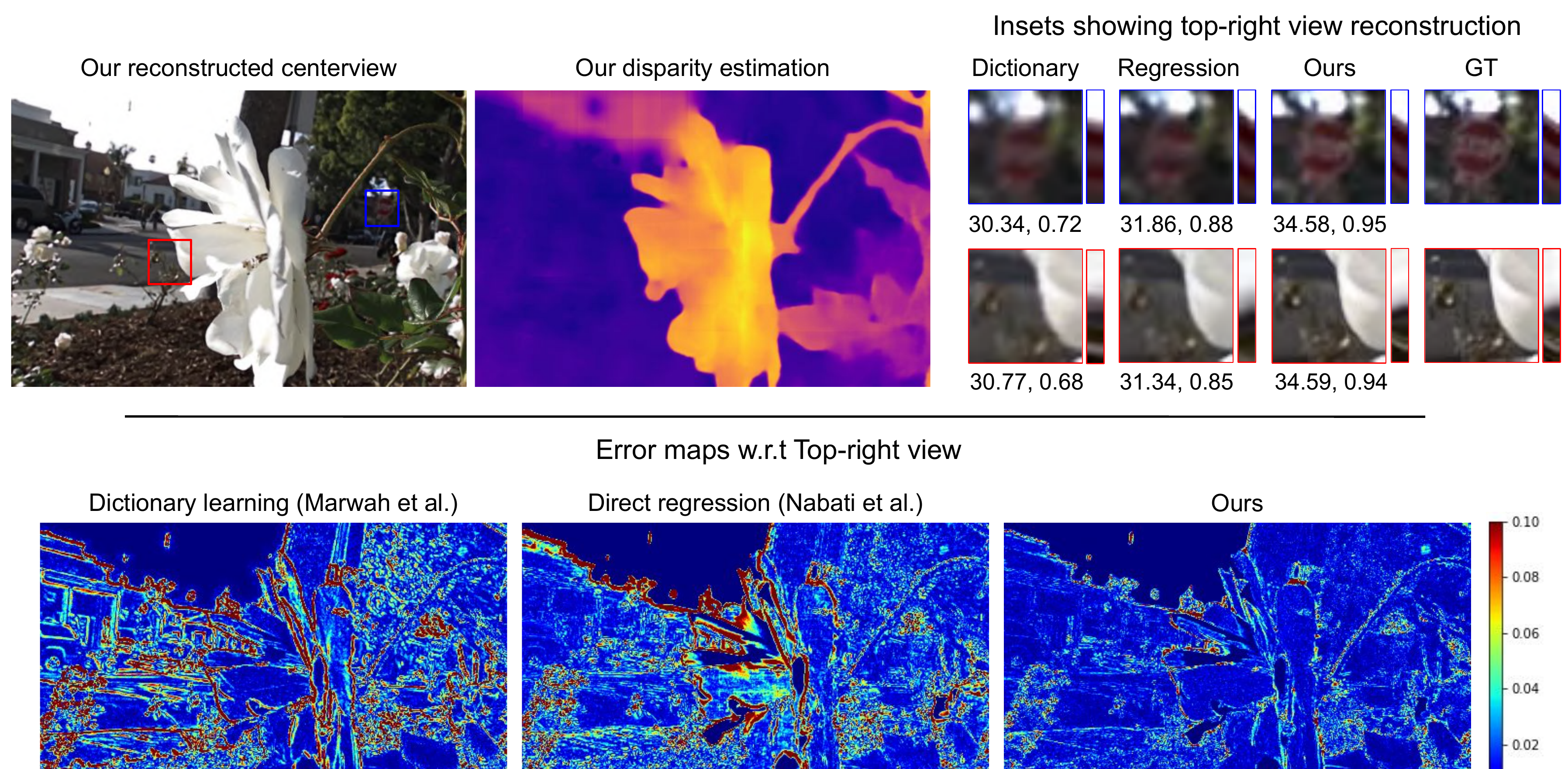}
    \caption{Compressive light field reconstruction $(5\times5\times320\times500)$ on Flowers2 test scene from Kalantari et al. test set: Top row shows our reconstructed centerview and the disparity map. In the insets, we show corresponding patches from the top-right view reconstruction along with EPIs. Below each inset, we provide the PSNR and SSIM values. Bottom row shows the error maps w.r.t the top-right view of the light field. Notice low error values in our reconstruction compared to dictionary learning and regression approaches especially around the occlusion edges of flowers. See Table \ref{tab:clf} for average results on testset.}
    \label{fig:clf_reconst}
\end{figure*}

\section{Training}
For training our framework we use the light field dataset\footnote{cseweb.ucsd.edu/~viscomp/projects/LF/papers/SIGASIA16/}  by Kalantari et al. consisting of 100 LFs for training, 30 for testing. In addition, we also show our results on Flowers dataset \footnote{github.com/pratulsrinivasan/Local\_Light\_Field\_Synthesis} by Srinivasan et al. with 3343 light field which we randomly split into 3100 for training and the rest for testing. We refer to these test LFs as Flowers testset. This data consists of macro-shots of flowers exhibiting complex occlusion patterns and weak textured regions like petals of the flowers. In both the datasets, the resolution of original LF is $14\times14\times375\times560$.  Since the angular views along the border exhibit artifacts we limit ourselves to angular resolution of $7\times7$.

We randomly crop $7\times7\times120\times120$ patches from the light fields for training our deep networks. We simulate the coded images from the light field and use it as input for our light field reconstruction. For warping, we use bilinear interpolation. We implemented all our models in Tensorflow and train the parameters by minimizing the overall loss function given in Eqn. \ref{eq:overall_loss}. Each convolutional layer is followed by exponential linear unit (ELU) and batch normalization (BN) except for the last layer in each network. The weights of the different regularizers are set to $\lambda_{dc} = 0.008$, $\lambda_{tv} = 0.01$. We use the Adam optimizer \cite{kingma2014adam} with a learning rate of $1e-4$, $\beta_1 = 0.9$, $\beta_2 = 0.999$, $\epsilon = 1e-8$ and a batch size of $7$. We train for a maximum of $100K$ iterations. Note that in all the three schemes for light field reconstruction - CLF, CA and focus-defocus pair, we use the same set of optimization parameters. 

\section{Results} \label{sec:results}
In all our multiplexing schemes, CLF, CA and FocDef pair, we synthesize light field of spatio-angular resolution $7\times7\times320\times500$ from coded images of spatial resolution $320\times500$. In the following section, we will discuss the light field reconstruction from coded images along with the comparisons for each multiplexing scheme. This is followed by ablation studies which show the effectiveness of using multiscale context aggregation and refinement of LF. Finally, we demonstrate real LF reconstructions with focus-defocus pairs captured from a DSLR. 

\subsection{Compressive Light Field Reconstrucion}
For simulating the coded image in this case, we sample code from a Gaussian distribution with values clipped to $[0,1$] and perform multiplexing similar to Marwah et al. \cite{marwah2013compressive}\footnote{web.media.mit.edu/~gordonw/CompressiveLightFieldPhotography/} for $7\times7$ LF. Note the code in this case is spatially varying and is of  $15\times15$ spatial resolution which we repeat to multiplex the whole image.

\begin{table}[t]
    \centering
    \caption{Comparison of CLF reconstruction for 5 test LFs ($5\times5\times320\times500$) from Kalantari et al. \cite{kalantari2016learning} dataset.}
    \begin{tabular}{c|c}
       Approach & 5 Test LFs \\
        \hline
    Dictionary \cite{marwah2013compressive}  & 32.46,	0.803 \\
    Regression \cite{nabati2018fast} & 33.77, 0.922 \\
    Ours CLF & \textbf{35.62}, \textbf{0.959}
    \end{tabular}
    \label{tab:clf}
\end{table}

\begin{figure*}[t]
    \centering
    \includegraphics[width=\textwidth]{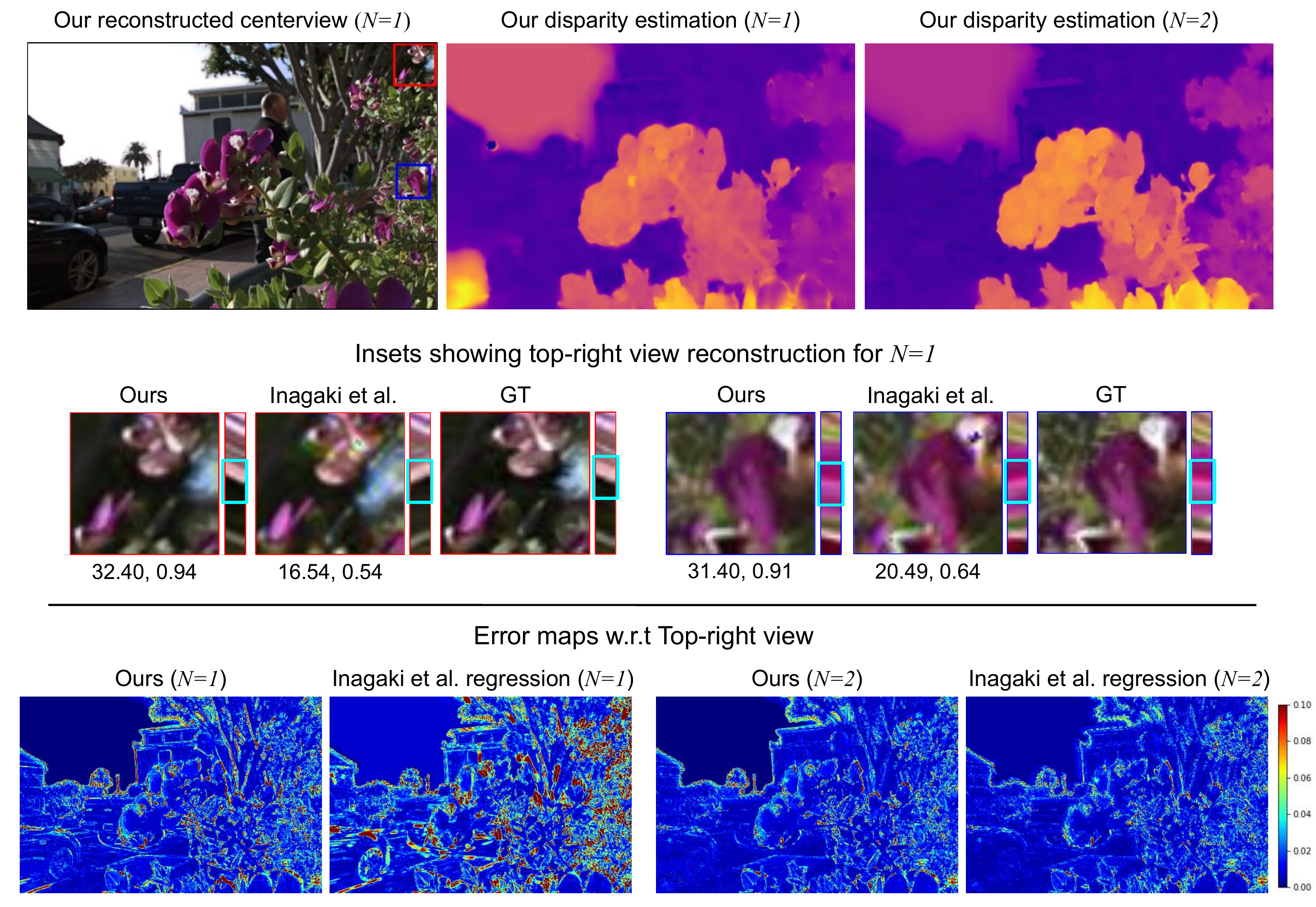}
    \caption{Coded aperture light field reconstruction $(7\times7\times320\times500)$ for Flower-1 test scene from Kalantari et al. test set: Top row shows our reconstructed centerview with one coded input (N=1) and the disparity maps with one (N=1) and two coded inputs (N=2), respectively. In the insets shown in blue and red, we show the corresponding patches (along with EPIs) from the top-right view of the reconstructed LF for $N=1$. PSNR and SSIM values are provided below. Notice that Inagaki et al. recover incorrect EPI slopes, which significantly affects their PSNR. Bottom row shows the error maps w.r.t the top-right view of the LF. Notice the low error values in our reconstructed top-right view in case of $N=1$, our reconstruction outperforms Inagaki et al. with minimal coded images as input. See Table \ref{tab:ca} for average results on testset.}
    \label{fig:ca_reconst}
\end{figure*}

Figure \ref{fig:clf_reconst} shows the results of our compressive LF reconstruction algorithm for the test scene, Flowers2 from the Kalantari et al. data. Our disparity map captures the scene geometry well by clearly delineating the flower from the background. 
Dictionary learning \cite{marwah2013compressive} does not recover the high frequency details well as can be seen from the details shown in the inset. Also, the EPI slopes are not recovered well which is essential for LF recovery thus affecting its PSNR and SSIM values. Direct regression \cite{nabati2018fast} reconstructs LF directly from the coded image without intermediate disparity. Although this approach reconstructs EPI slopes better than dictionary learning it still does not recover the high frequency details as well as our approach. In contrast to dictionary learning and direct regression, our approach does LF reconstruction in two stages of view-reconstruction and disparity estimation. By delineating the LF reconstruction into these two stages, our approach recovers light field better than the other methods as indicated in Table \ref{tab:clf} and the error maps in Fig. \ref{fig:clf_reconst}. 

Table \ref{tab:clf} shows the average LF reconstruction results on 5 test LFs from Kalantari et al. dataset. We train light field dictionaries similar to Marwah et al. using a subset of training LFs from Kalantari et al. dataset. For reconstruction using dictionaries, we solve LASSO with ADMM. For direct regression approach evaluation, we trained the architecture of Nabati et al. \cite{nabati2018fast} similar to that of our approach.

\subsection{Coded Aperture Light Field Reconstruction} For simulating the coded image, we sample code from a uniform distribution in $[0,1$] range. Note the code in this case is spatially invariant (see Fig. \ref{fig:teaser}) and hence multiplexes only the angular dimensions of $7\times7$ resolution. Compared to compressive light field, the coded aperture measurements are less incoherent and hence both view reconstruction and disparity estimation are challenging. 

Figure \ref{fig:ca_reconst} shows the results of our coded aperture LF reconstruction for the test scene, Flower-1 from Kalantari et al. data. We show our reconstructed centerview from the coded image along with the disparity maps. We compare our results with the learning approach of Inagaki et al. \cite{inagaki2018learning} which reconstructs LF from coded images directly without any disparity estimation. For comparison, we use the coded aperture masks provided by Inagaki et al. for the case of single $(N=1)$ and dual $(N=2)$ coded aperture at $8\times8$ angular resolution and resize them to $7\times7$. Using these masks, we train both ours and Inagaki et al. approach on Kalantari et al. dataset. Table \ref{tab:ca} shows the average reconstruction results over the testset. 

As we can see from the error maps shown in the figure and Table \ref{tab:ca}, for the challenging case of single coded image $(N=1)$, our approach outperforms Inagaki et al. \cite{inagaki2018learning} by 2 dB on the testset. The red and blue insets shown in the figure detail the top-right view reconstructed along with the EPIs for $N=1$ case. We can see that Inagaki et al. reconstruction has artifacts as evident in the blue inset. Also, it incorrectly estimates the EPI slopes, see the highlighted regions on the EPI for both the insets. For $N=2$ case, both the approaches are comparable as can be seen from error maps in Figure \ref{fig:ca_reconst} and the average test results in Table \ref{tab:ca}. 

\begin{table}[t]
    \centering
    \caption{Coded aperture reconstruction results on 30 test LFs ($7\times7\times320\times500$) from Kalantari et al. \cite{kalantari2016learning} dataset}
    \begin{tabular}{c|c|c}
         &  $N = 1$ & $N=2$ \\
        \hline
    Inagaki et al. \cite{inagaki2018learning} & 34.71, 0.85 & \textbf{39.23}, 0.93 \\
        Ours & \textbf{37.25, 0.94} & 38.80, \textbf{0.96}
    \end{tabular}
    \label{tab:ca}
\end{table}

\subsection{Light Field from Focus-Defocus Pair}
For simulating defocus image, we take the average across all the views ($7\times7$) which amounts to an $f$ number of $f/4$. We additionally augment the data by shearing the light field. This shearing results in shifting of the focal plane in the defocus image, thus ensuring that during training, disparity estimation network sees defocus at various depths relative to the focal plane. Since we provide the centerview explicitly as all-in-focus scene, view reconstruction is not necessary in this case.

\begin{figure}[t]
    \centering
    \centerline{$l_1$ error vs. views of light field}
    \includegraphics[trim={0 0 0 0},clip,width=0.43\textwidth]{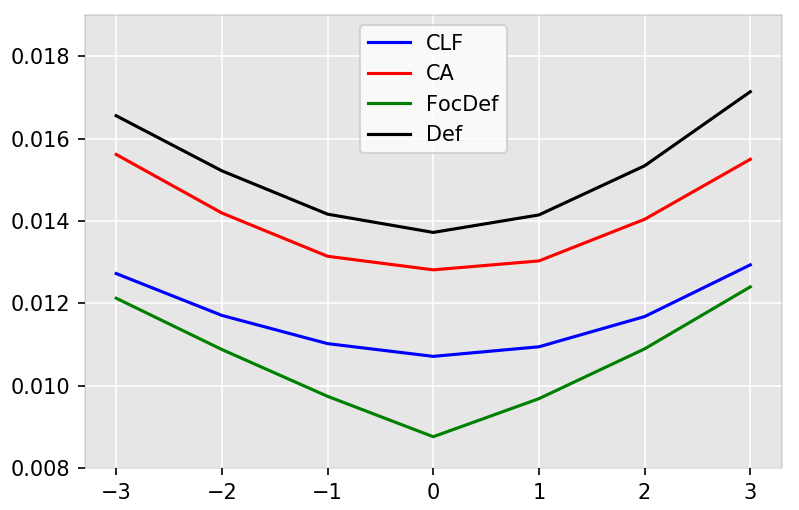}
    \caption{Comparison of test error ($l_1$) across views over testset of Kalantari et al. with 30 LFs. Views are averaged over along the vertical angular dimension. Comparison is shown for compressive LF (CLF), coded aperture (CA), focus-defocus (FocDef) and only defocus (Def).}
    \label{fig:l1err_cmp_frameworks}
\end{figure}

Note that one can try to reconstruct LF from a single defocused image also. But, compared to CLF and CA, this is a severely ill-posed task as a defocused image corresponds to uniform angular multiplexing of the LF. 
Figure \ref{fig:l1err_cmp_frameworks} shows the LF reconstruction error for different multiplexing schemes. As can be seen, the LF reconstruction from just a single defocused image performs poorly compared to CLF and CA. To compensate for the loss of angular information in the defocused image, we propose to use the  all-in-focus image as well for LF reconstruction. With this addition, we get our focus-defocus LF reconstruction scheme, whose performance is even better than CLF, see Fig. 
\ref{fig:l1err_cmp_frameworks}. Moreover, the addition of the in-focus image does not involve any hardware change to the conventional camera.

\begin{figure*}[t]
    \centering
    \begin{minipage}{0.18\textwidth}    \centering    Input defocus    \end{minipage}
    \begin{minipage}{0.155\textwidth}    \centering    Single image \\ Srinivasan et al.  \end{minipage}
    \begin{minipage}{0.155\textwidth}    \centering    View synthesis \\ Kalantari et al. \end{minipage}
    \begin{minipage}{0.155\textwidth}    \centering    Ours CLF  \end{minipage}
    \begin{minipage}{0.155\textwidth}    \centering    Ours Foc-Def     \end{minipage}
    \begin{minipage}{0.155\textwidth}    \centering    From 4D LF \\ Shin et al.    \end{minipage}
    \\\vspace{0.05cm}
    \IText{\includegraphics[width=.155\textwidth]{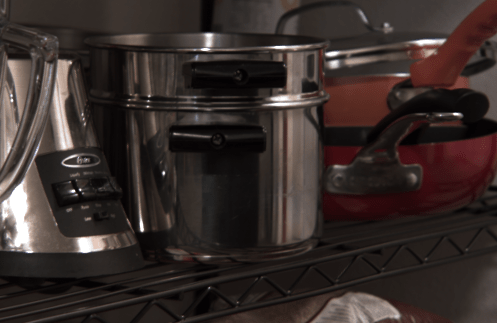}}{\small Reflective}
    \includegraphics[width=.155\textwidth]{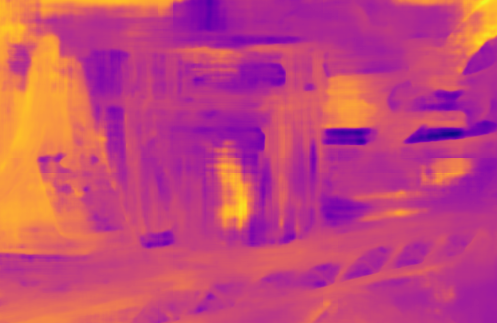}
    \includegraphics[width=.155\textwidth]{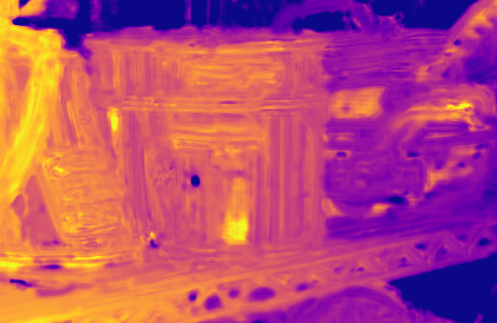}
    \includegraphics[width=.155\textwidth]{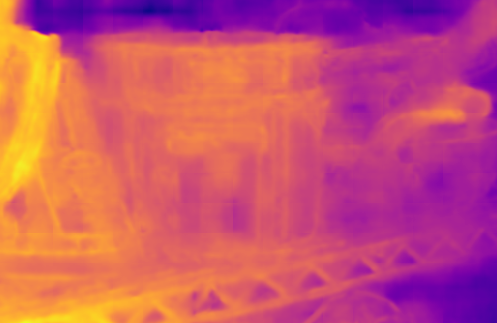}
    \includegraphics[width=.155\textwidth]{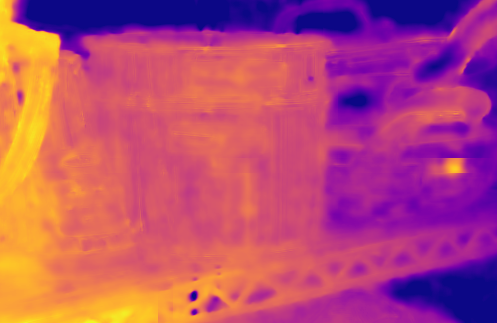}
    \includegraphics[width=.155\textwidth]{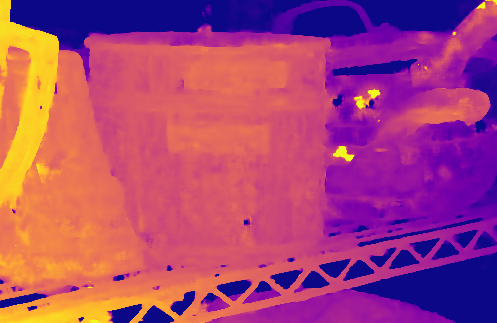}
    \\\vspace{0.05cm}
    \IText{\includegraphics[width=.155\textwidth]{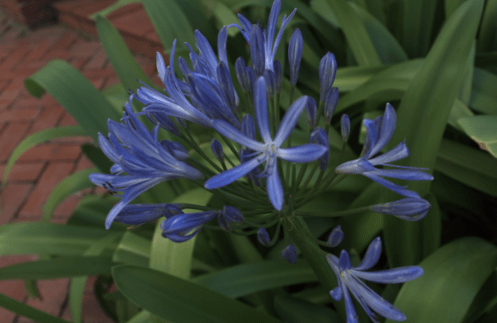}}{\small Flower-3}
    \includegraphics[width=.155\textwidth]{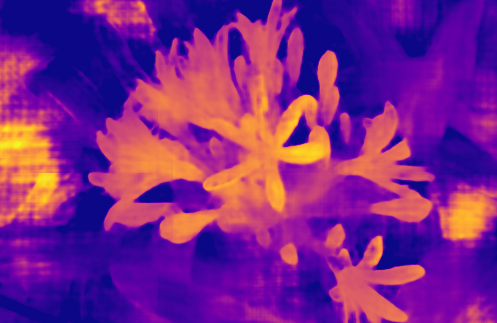}
    \includegraphics[width=.155\textwidth]{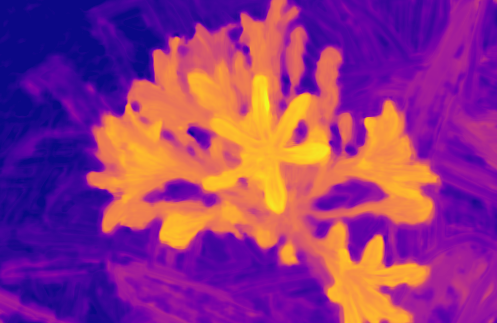}
    \includegraphics[width=.155\textwidth]{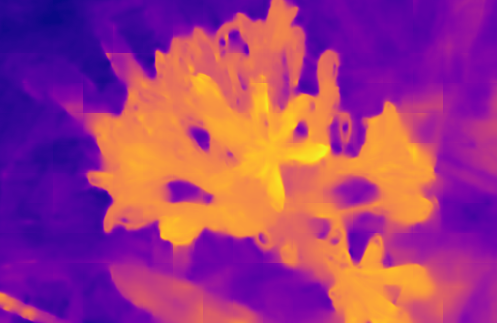}
    \includegraphics[width=.155\textwidth]{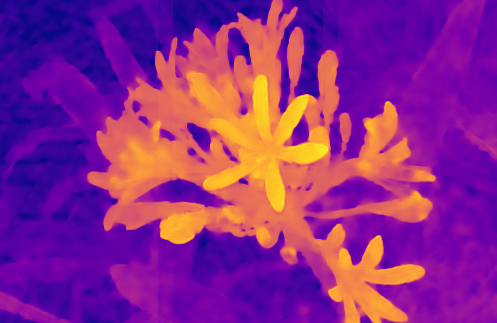}
    \includegraphics[width=.155\textwidth]{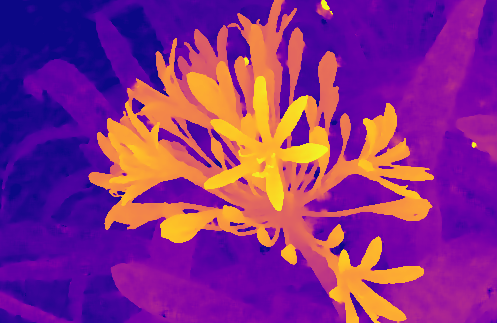}
    \\\vspace{0.05cm}
    \IText{\includegraphics[width=.155\textwidth]{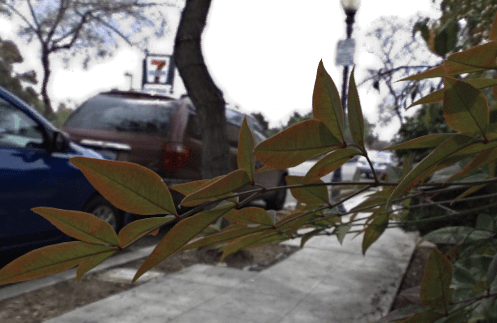}}{\small Leaves}
    \includegraphics[width=.155\textwidth]{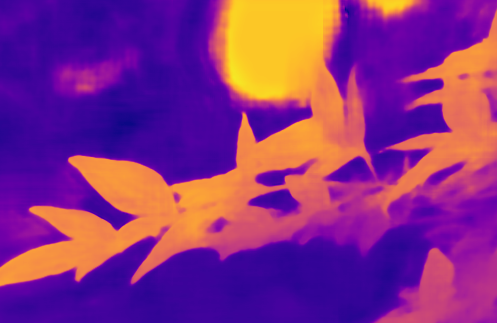}
    \includegraphics[width=.155\textwidth]{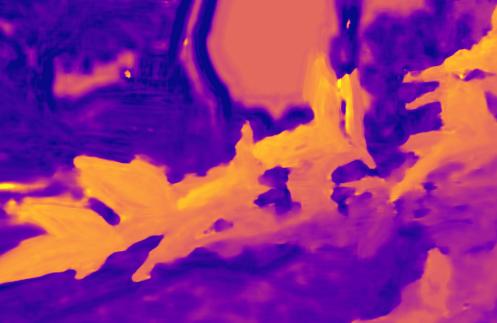}
    \includegraphics[width=.155\textwidth]{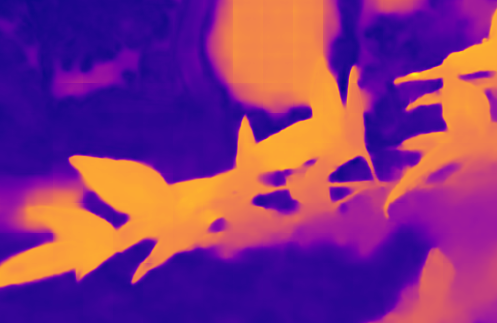}
    \includegraphics[width=.155\textwidth]{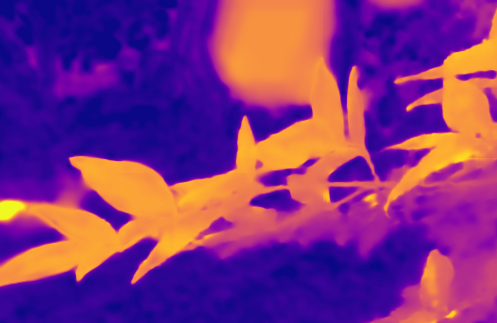}
    \includegraphics[width=.155\textwidth]{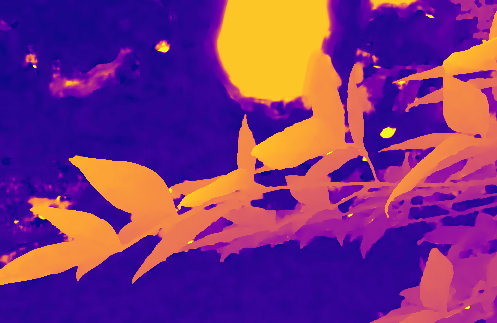}
    \\\vspace{0.05cm}
    \IText{\includegraphics[width=.155\textwidth]{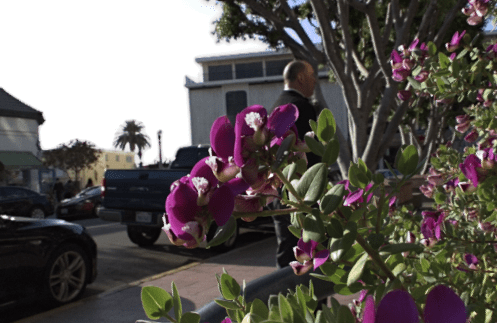}}{\small Flower-1}
    \includegraphics[width=.155\textwidth]{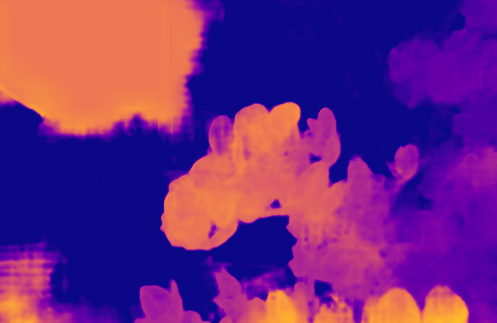}
    \includegraphics[width=.155\textwidth]{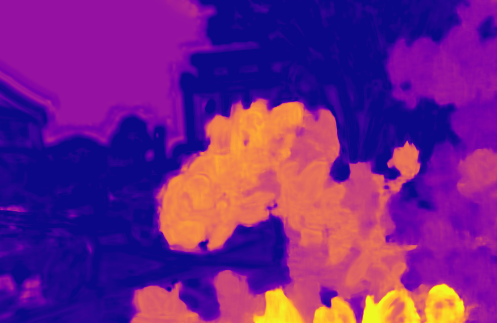}
    \includegraphics[width=.155\textwidth]{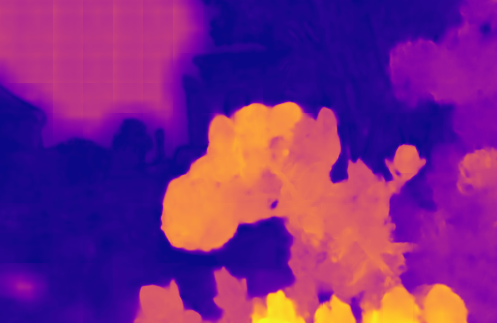}
    \includegraphics[width=.155\textwidth]{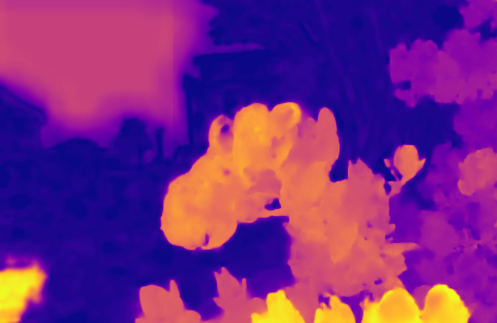}
    \includegraphics[width=.155\textwidth]{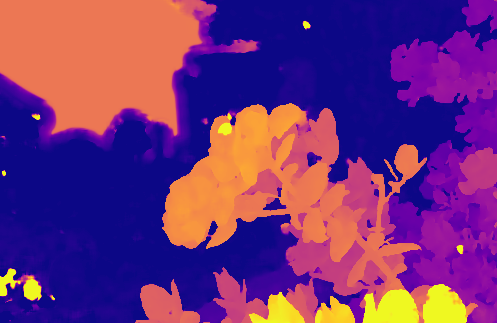}
    \caption{Disparity estimation comparison of ours Compressive LF and focus-defocus pair with single image method of \cite{srinivasan2017learning}, four corner views of \cite{kalantari2016learning} and  full 4D light field method of \cite{shin2018epinet}. The scenes are `Reflective' from Stanford Lytro LF archive \cite{stanfordlytroLF2016} and `Flower-3', `Leaves' and `Flower-1' from TestSet of Kalantari et al. As we can see, the single image method of \cite{srinivasan2017learning} lacks generalization across scenes. Our focus-defocus method contains sharper occlusion edges than our CLF method. Also, our disparities from focus-defocus pairs are slightly better than the multi-view estimation of Kalanatari et al. In `Leaves' and `Flower-3' scenes, the disparities obtained from the multi-view method bleed around occlusion edges whereas our estimation retains their sharpness (see Fig. \ref{fig:kal_cmp_disp} for detailed comparison).} 
    \label{fig:disp_cmp_others}
\end{figure*}

Since there exists no prior work with LF reconstruction from focus-defocus pairs, we compare our LF reconstruction with the recent state-of-the-art learning based LF synthesis approaches of Kalantari et al. \cite{kalantari2016learning} from four corner views and Srinivasan et al. \cite{srinivasan2017learning} from a single image. Figure \ref{fig:disp_cmp_others} and Table \ref{tab:psnr_cmp_others} show comparisons with these approaches. For Kalantari et al. approach we use the code and trained models provided by the authors. For Srinivasan et al. approach we use the code made online by the authors for training and testing.

Both the approaches of Kalantari et al. and Srinivasan et al. involve disparity based LF synthesis. Figure \ref{fig:disp_cmp_others} compares the centerview disparity estimated from these approaches and ours with that of the disparity map obtained from full 4D LF by EPINET \cite{shin2018epinet}. EPINET is a deep network operating along EPIs of light field for disparity estimation. As we can see from the disparities, single image based approach, being trained on the specific dataset of flowers, fails to generalize across all scenes. For scenes like Leaves and Flower-1 it performs well, but for others scenes such as Flower-3, Reflective (Fig. \ref{fig:disp_cmp_others}) and Seahorse (Fig. \ref{fig:teaser}), it fails.

\begin{table}[t]
    \caption{Comparison of different light field reconstruction ($7\times7\times320\times500$) algorithms on test sets of Kalantari et al. \cite{kalantari2016learning} and Flowers dataset \cite{srinivasan2018aperture}. Except the single image approach of Srinivasan et al. \cite{srinivasan2017learning} rest of the approaches are trained on the respective training sets. Single image approach is trained on Flowers dataset alone as it does not generalize on the Kalantari et al. dataset due to small number of LFs. }
    \label{tab:psnr_cmp_others}
    \begin{minipage}{\columnwidth}
    \begin{center}
    \begin{tabular}{l|c|c}
         & Kalantari et al. & Flowers  \\
        Approach &  TestSet (30) & TestSet (100) \\
        \hline
        Single image  \cite{srinivasan2018aperture} (N=1) & 35.55, 0.930 & 37.88, 0.935 \\
        Compressive LF (N=1) & 37.29, 0.951 & N/A \\
        Coded aperture LF (N=1) & 36.84, 0.938 & N/A \\
        Defocus (N=1) & 36.42, 0.929 & N/A \\
        Focus-Defocus (N=2) & \textbf{38.21, 0.957} & 38.19, \textbf{0.946} \\
        View synthesis  \cite{kalantari2016learning} (N=4) & 37.51, 0.956  &  \textbf{38.31}, 0.938 
    \end{tabular}   
    \end{center}
    \end{minipage}
\end{table}

\begin{figure}[t]
    \centering
    \includegraphics[width=0.37\textwidth]{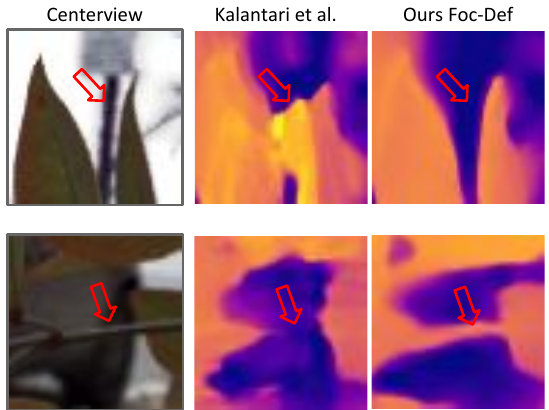}
    \caption{Comparison of disparities with view synthesis approach of Kalantari et al. \cite{kalantari2016learning}. Notice how our approach captures fine details that are missed by the view synthesis approach as pointed by the red arrow.}
    \label{fig:kal_cmp_disp}
\end{figure}

LF synthesis from four corner views by Kalantari et al. uses a four layer CNN for disparity estimation with explicit stereo and defocus features as input. It results in better disparity maps but due to a rather simple CNN architecture it results in texture leakage and  bleeding around depth edges and misses out fine details sometimes in disparity maps. For example, for Seahorse scene (Fig. \ref{fig:teaser}) and  Reflective scene (Fig. \ref{fig:disp_cmp_others}), disparities from corner views exhibit texture leakage which are not prominent in our disparities estimated using focus-defocus input. In Fig. \ref{fig:kal_cmp_disp}, we compare our estimated disparities with Kalantari et al., as you can see, our estimation has lesser artifacts and preserves the fine details. 
Our approach performs well even for scenes like Reflective (Fig. \ref{fig:disp_cmp_others}) where the jar handle in the front is separated from the background, which is mostly transparent.

Table \ref{tab:psnr_cmp_others} shows the average results on testsets of Kalantari et al. \cite{kalantari2016learning} dataset and Flowers dataset of Srinivasan et al. \cite{srinivasan2018aperture}. The average values are reported by considering views which are not given as input to any of the methods. All the approaches are trained on respective datasets except the single image approach as it does not generalize on the Kalantari et al. training set of 100 LFs. So we train this on Flowers dataset alone and evaluate on both testsets. It performs poorly on Kalantari et al. testset whereas on Flowers testset it is relatively better. Focus-defocus LF reconstruction performs as well as the view synthesis approach of Kalantari et al. \cite{kalantari2016learning} on both the testsets. As expected, it performs better than the single image based LF reconstruction as it uses the defocus cue for disparity estimation whereas single in-focus image does not have any such cues. In Table  \ref{tab:psnr_cmp_others}, we also show the performance of CLF and CA on Kalantari et al. testset. Note that for simulating coded images for both CLF and CA, we used random masks as discussed earlier.

\subsection{Additional results on the Flowers dataset}
\begin{figure*}[t]
    \centering
    \begin{minipage}{0.24\textwidth}    \centering     In-focus image    \end{minipage}
    \begin{minipage}{0.24\textwidth}    \centering   Our disparity
    \end{minipage}
    \begin{minipage}{0.24\textwidth}    \centering     In-focus image    \end{minipage}
    \begin{minipage}{0.24\textwidth}    \centering   Our disparity
    \end{minipage}
    \\\vspace{0.1cm}
    \includegraphics[width=0.24\textwidth]{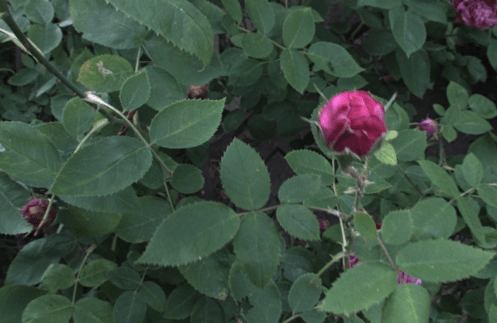}
    \includegraphics[width=0.24\textwidth]{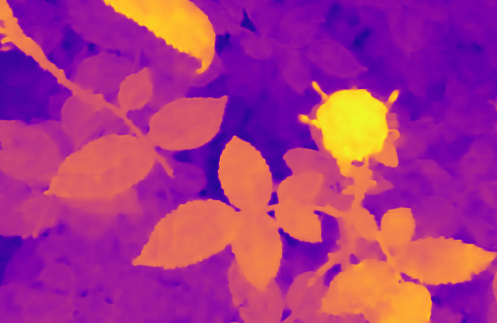}
    \includegraphics[width=0.24\textwidth]{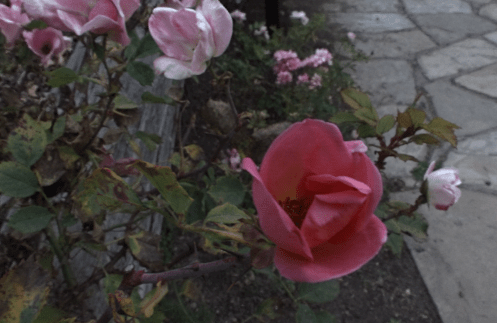}
    \includegraphics[width=0.24\textwidth]{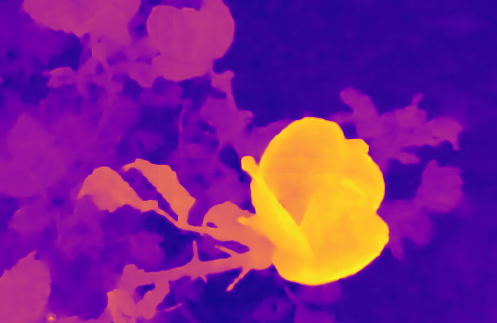}
    \\\vspace{0.08cm}
    \includegraphics[width=0.24\textwidth]{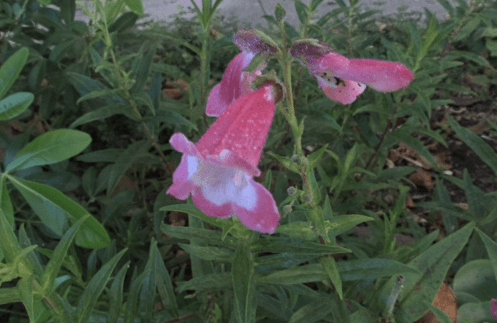}
    \includegraphics[width=0.24\textwidth]{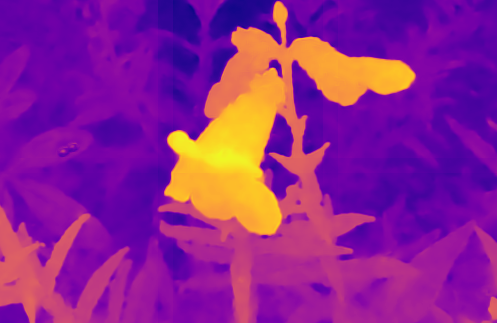}
    \includegraphics[width=0.24\textwidth]{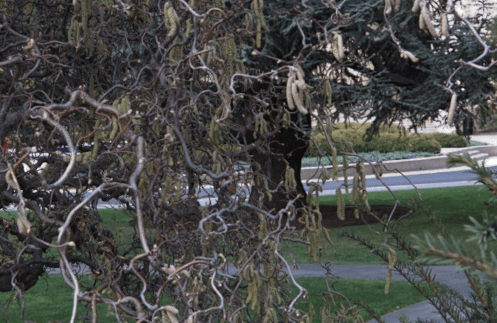}
    \includegraphics[width=0.24\textwidth]{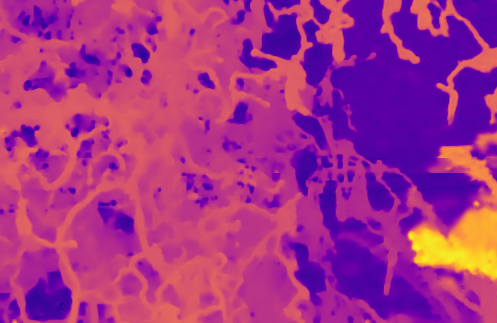}
    \caption{Disparity maps using focus-defocus input on test images of Flowers dataset. First three scenes are from Flowers testset, the bottom right scene ('Occlusions\_16') is from the Stanford LF archive  \cite{stanfordlytroLF2016}. Notice the fine details and occlusion patterns captured in our disparity maps.}
    \label{fig:disp_singfl_cmp}
\end{figure*}

Figure \ref{fig:disp_singfl_cmp} shows the disparity maps estimated using the focus-defocus input for Flowers dataset. Scenes in this data exhibit challenges like constant intensity regions and complex occlusion patterns. As can be seen from the disparity maps our approach captures the fine occlusion information created by thin stems and plant leaves. Also, it estimates the disparity values accurately for flower petals which have constant intensity values. This is due to our multiscale feature aggregation step in our DisparityNet. As we show in our ablation study in the next section, without that, our estimates are inaccurate for such challenging scenes. Average results of LF reconstruction are provided in Table \ref{tab:psnr_cmp_others}.  


\subsection{Multiscale context aggregation} 
\begin{figure*}[t]
    \centering
    \begin{minipage}{.2\textwidth}  \centering Scene \end{minipage}
    \begin{minipage}{.2\textwidth}  \centering Without aggregation \end{minipage}
    \begin{minipage}{.2\textwidth}  \centering With aggregation \end{minipage}
    \begin{minipage}{.2\textwidth}  \centering From 4D LF \cite{shin2018epinet} \end{minipage}
    \\\vspace{0.1cm}
    \includegraphics[width=0.2\textwidth]{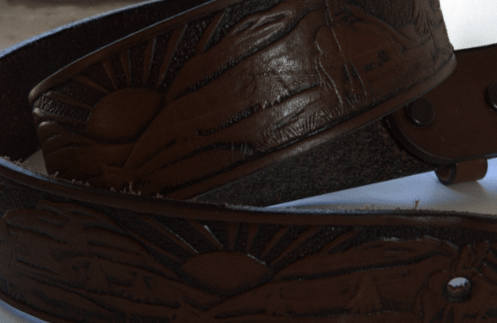}
    \includegraphics[width=0.2\textwidth]{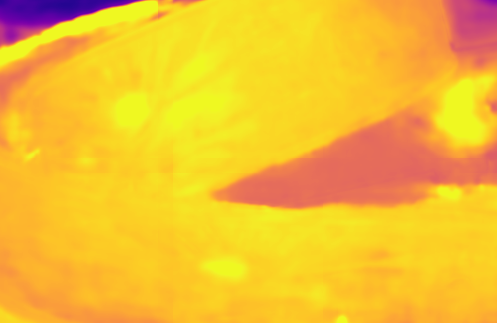}
    \includegraphics[width=0.2\textwidth]{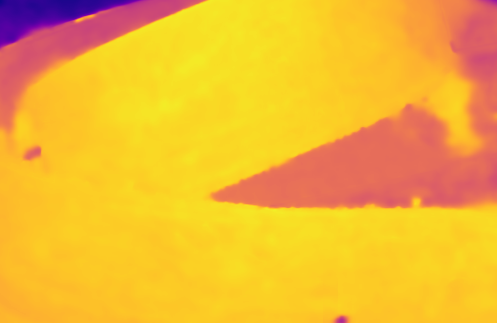}
    \includegraphics[width=0.2\textwidth]{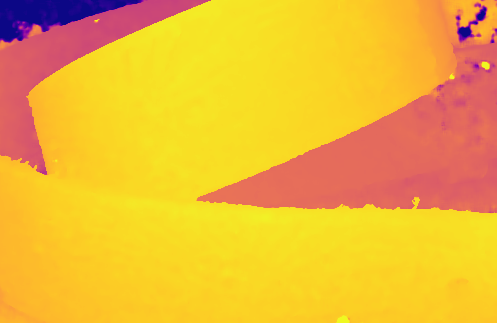}
    \\\vspace{0.1cm}
    \includegraphics[width=0.2\textwidth]{figures/dispest/6code.png}
    \includegraphics[width=0.2\textwidth]{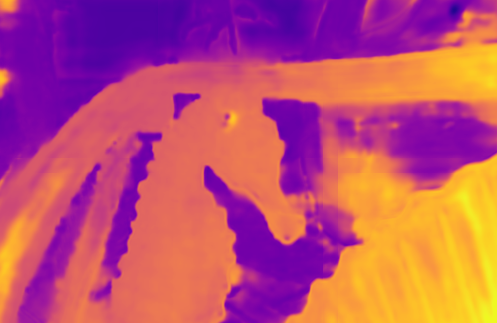}
    \includegraphics[width=0.2\textwidth]{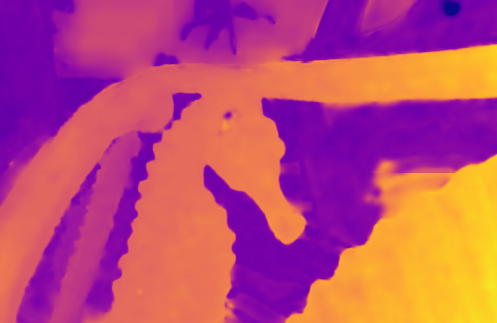}
    \includegraphics[width=0.2\textwidth]{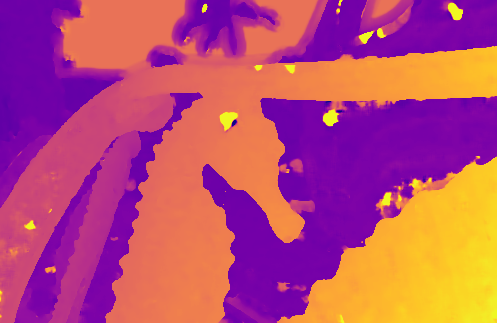}
    \caption{Effect of disparity estimation with and without multiscale feature aggregation shown in case of focus-defocus LF reconstruction. Figure shows the centerview and the corresponding disparity maps. Notice the texture leakage for the belt scene without the aggregation which is suppressed by using aggregation. Also, without aggregation, the disparity is incorrect for the latter edge of the belt, whereas with aggregation it is closer to the disparity obtained from full LF by Shin et al.  \cite{shin2018epinet}. Further, notice the inconsistent disparity estimation by without aggregation for constant brightness regions like car body in the Seahorse scene.}
    \label{fig:noMS}
\end{figure*}

\textcolor{black}{Without the context aggregation block, our DisparityNet is similar to the disparity estimation architecture proposed for single image based LF reconstruction by Srinivasan et al. \cite{srinivasan2018aperture}. Note that when using Srinivasan et al. architecture we modify their input layer for our input focus-defocus images and retrain it. Figure \ref{fig:noMS} shows the disparity estimation without the multiscale feature aggregation in DisparityNet (see Sec. \ref{fig:dispnet}). As can seen from the figure, our disparity estimation is qualitatively better in weakly textured regions (car surface) and exhibits less RGB texture leakage (on the belt). On the testset of Kalantari et al. consisting of 30 LFs, our network without context aggregation achieves PSNR and SSIM of $37.51$ dB, $0.944$ and with context aggregation, achieves $38.21$ dB, $0.957$.} 

\begin{figure}[t]
\includegraphics[width=0.45\textwidth]{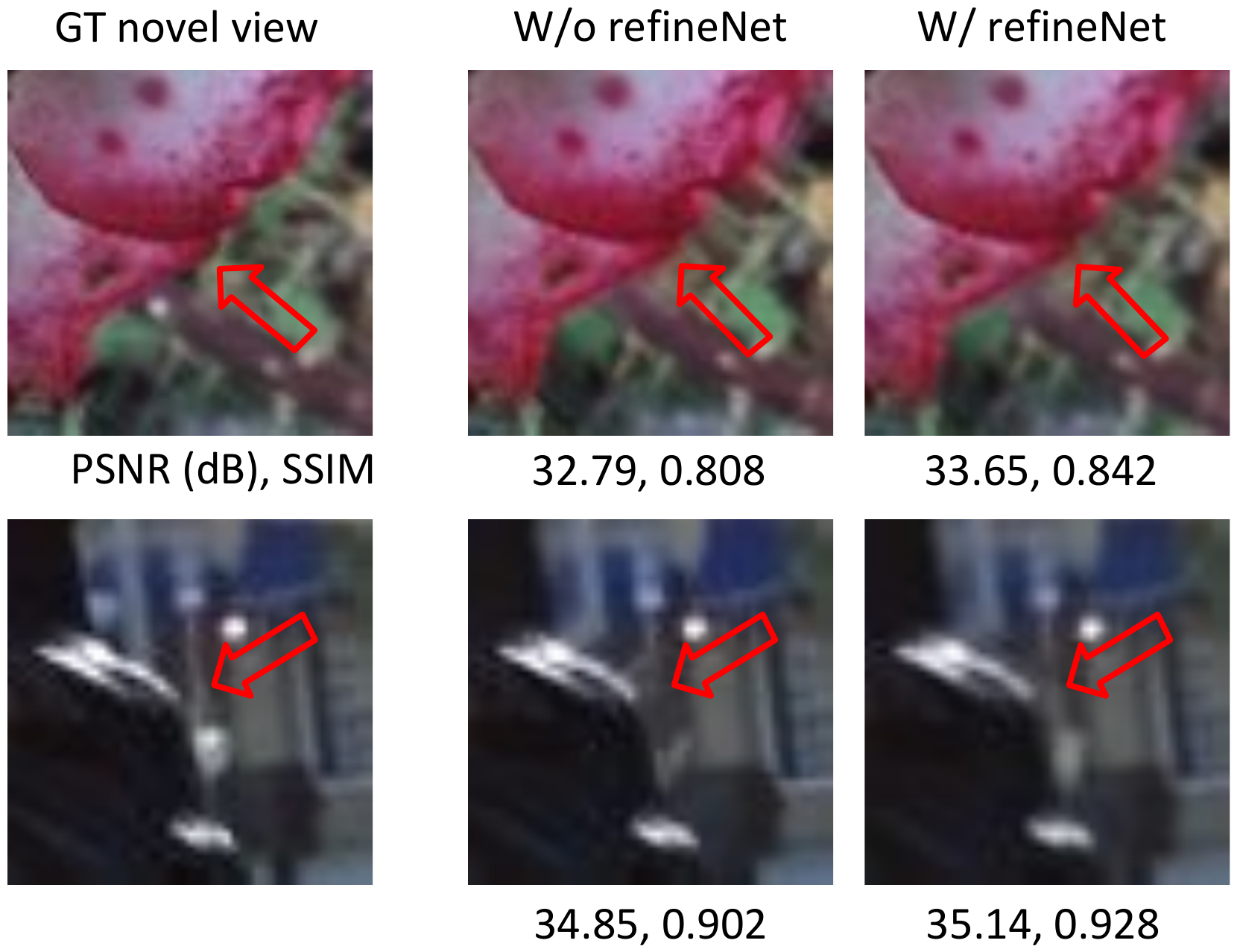}
\caption{\textcolor{black}{Light field view reconstruction without and with the refinement stage. Notice the suppression of shearing artifacts around occlusion edges highlighted by the red arrow and improvement in PSNR, SSIM.}}
\label{fig:refine_final}
\end{figure}

\subsection{Light field refinement}
Figure \ref{fig:refine_final} shows the LF reconstruction results before and after the refinement stage. Warping based LF reconstruction introduces shearing artifacts around occlusion edges. Refinement stage takes as input the warped LF and the disparity field and predicts a residual LF which is then added to the warped LF. The refined LF exhibits less artifacts around occlusion edges as seen in the Fig. \ref{fig:refine_final}.  On the Flowers test of 100 LFs, light field reconstruction from focus-defocus without and with refinement gave PSNR and SSIM of  38.31 dB, 0.944 and 38.95 dB, 0.954, respectively.

\subsection{Comparison with encoder-decoder architecture}
We compare our dilated convolution and multiscale context aggregation based \textit{DisparityNet} with the encoder-decoder architure based approach of Vadathya et al. \cite{vadathya2018learning}. We show this comparison for our best performing case of focus-defocus scheme. Vadathya et al. use encoder-decoder architecture for disparity estimation with skip connections to preserve the low-level details. We use dilated convolution to enhance receptive field and perform multiscale context aggregation to resolve ambiguities in disparity estimation. On the Kalantari et al. testset of 30 LFs our network achieves 38.21dB (PSNR), 0.96 (SSIM) whereas the encoder-decoder architecture results in 37.28 dB, 0.94.

\subsection{LF synthesis from a DSLR}
To demonstrate real light field reconstructions, we capture in-focus and defocus pair from a DSLR (Canon 70D) by varying the aperture (f/4, f/16). Figure \ref{fig:real_lf} shows the LF reconstruction ($7\times7\times460\times610$) results with focus-defocus pair captured using a conventional camera. Notice the relative disparity estimation (+ve and -ve) delineating the depth planes.

\section{Discussions and Conclusion}
We propose a unified learning framework for \textit{full sensor resolution} light field reconstruction which can work under a \textit{variety of spatio-angular multiplexing schemes with minimal measurements}. We demonstrate this on three coded schemes: compressive LF with code near the sensor (CLF), coded aperture (CA) with code on the aperture and focus-defocus pair from a conventional camera. If we consider single shot imaging, then CLF followed by CA are better posed than just a defocused image as they do explicit coding of spatio-angular rays. However, by adding an all-in-focus image (captured with narrow aperture), along with the defocused image, the reconstruction problem becomes much better posed as we can use the fine texture details from the all-focus image and depth cue from relative defocus between the two images. The focus-defocus scheme has the additional advantage of ease of capture using conventional cameras.  

\begin{figure*}[!t]
    \centering
    \includegraphics[trim={0 2.9in 0 0}, clip, width=.98\textwidth]{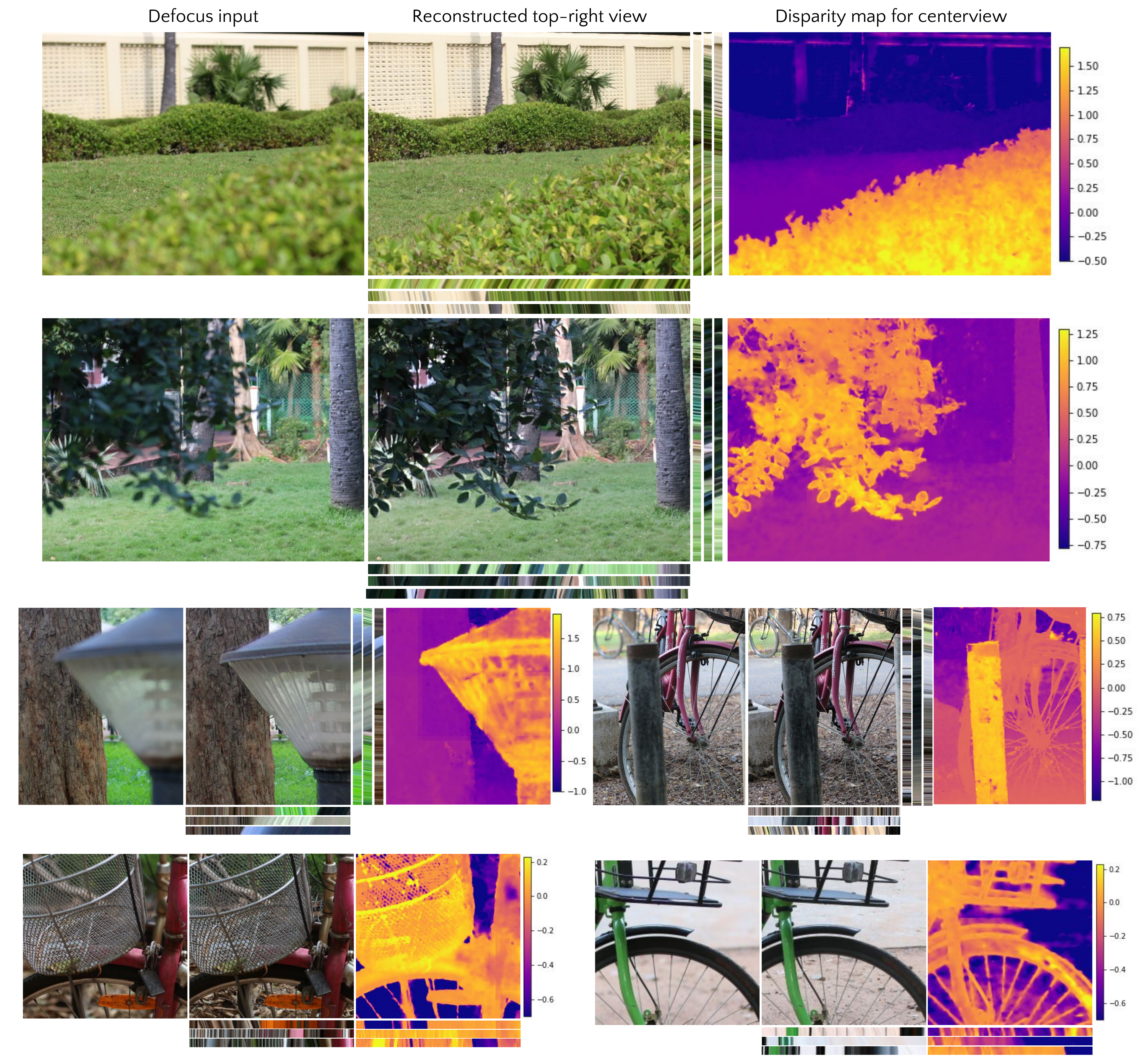}
    \caption{Real LF reconstruction results from a DSLR, where the focus-defocus images are obtained by camera apertures of f/16 and f/4 respectively. Each scene shows the defocus image, top-right reconstructed view overlaid with EPI and centerview disparity map.}
    \label{fig:real_lf}
\end{figure*}

Using our disparity based LF reconstruction framework, we analyze these three coded imaging schemes in terms of \textit{view reconstruction} and \textit{disparity estimation}.  Disparity estimation is learned in an unsupervised manner by using the light field reconstruction loss. As can be seen in our results, disparity maps estimated from focus-defocus pair are better than that from CLF and are similar to those obtained from full 4D light field (see Fig. \ref{fig:disp_cmp_others}). Focus-defocus utilizes the high frequency details from the all-in focus image and hence results in sharper disparities as compared to CLF. Focus-defocus also recovers the relative disparities for depth planes in-front and behind the focal planes very well (see +ve and -ve disparities in Fig. \ref{fig:real_lf}). This delineation is challenging as local cue such as relative defocus is not enough and we need to utilize global cue such as occlusion edges. The feature aggregation step in our disparity estimation network leverages on this global cue. In addition, we also have a data-augmentation step where shearing the light field changes the focal plane. This ensures that the data has variety of relative defocus w.r.t the focal plane during training.

We compared our reconstructions with both direct regression (Nabati et al. \cite{nabati2018fast}, Inagaki et al. \cite{inagaki2018learning}) and disparity based rendering from four corner views of LF (Kalantari et al. \cite{kalantari2016learning}) and from a single image (Srinivasan et al. \cite{srinivasan2018aperture}). Our CLF and focus-defocus approaches perform as well as the four corner view method of Kalantari et al and outperform  direct regression and single image based approaches. Moreover, our approach is much faster than that of Kalantari et al. In the four view approach the disparities are estimated independently for each view, which is computationally inefficient, whereas in our approach we compute disparities for all the views jointly. 

Our learning algorithm, thus, paves the way for full sensor resolution light field reconstruction from either a single coded image with specialized hardware (CLF and CA) or with two images (FocDef) from a conventional camera. 

%



\section*{Acknowledgment}
This work is supported by Qualcomm Innovation Fellowship (QInF) India 2016 and 2017. 


\ifCLASSOPTIONcaptionsoff
  \newpage
\fi



%

\bibliographystyle{IEEEbib}
\bibliography{refs}

\begin{thebibliography}{10}

\bibitem{schedl2015directional}
David~C Schedl, Clemens Birklbauer, and Oliver Bimber,
\newblock ``Directional super-resolution by means of coded sampling and guided
  upsampling,''
\newblock in {\em ICCP}. IEEE, 2015, pp. 1--10.

\bibitem{mitra2012light}
Kaushik Mitra and Ashok Veeraraghavan,
\newblock ``Light field denoising, light field superresolution and stereo
  camera based refocussing using a gmm light field patch prior,''
\newblock in {\em CVPRW}. IEEE, 2012, pp. 22--28.

\bibitem{shi2014light}
Lixin Shi, Haitham Hassanieh, Abe Davis, Dina Katabi, and Fredo Durand,
\newblock ``Light field reconstruction using sparsity in the continuous fourier
  domain,''
\newblock {\em ACM TOG}, vol. 34, no. 1, pp. 12, 2014.

\bibitem{wanner2014variational}
Sven Wanner and Bastian Goldluecke,
\newblock ``Variational light field analysis for disparity estimation and
  super-resolution,''
\newblock {\em Trans. on PAMI}, vol. 36, no. 3, pp. 606--619, 2014.

\bibitem{georgiev2006spatio}
Todor Georgiev, Ke~Colin Zheng, Brian Curless, David Salesin, Shree Nayar, and
  Chintan Intwala,
\newblock ``Spatio-angular resolution tradeoffs in integral photography.,''
\newblock {\em Rendering Techniques}, vol. 2006, pp. 263--272, 2006.

\bibitem{kalantari2016learning}
Nima~Khademi Kalantari, Ting-Chun Wang, and Ravi Ramamoorthi,
\newblock ``Learning-based view synthesis for light field cameras,''
\newblock {\em ACM TOG}, vol. 35, no. 6, pp. 193, 2016.

\bibitem{ashok2010compressive}
Amit Ashok and Mark~A Neifeld,
\newblock ``Compressive light field imaging,''
\newblock in {\em Three-Dimensional Imaging, Visualization, and Display 2010
  and Display Technologies and Applications for Defense, Security, and Avionics
  IV}. International Society for Optics and Photonics, 2010, vol. 7690, p.
  76900Q.

\bibitem{marwah2013compressive}
Kshitij Marwah, Gordon Wetzstein, Yosuke Bando, and Ramesh Raskar,
\newblock ``Compressive light field photography using overcomplete dictionaries
  and optimized projections,''
\newblock {\em ACM TOG}, vol. 32, no. 4, pp. 46, 2013.

\bibitem{liang2008programmable}
Chia-Kai Liang, Tai-Hsu Lin, Bing-Yi Wong, Chi Liu, and Homer~H Chen,
\newblock ``Programmable aperture photography: multiplexed light field
  acquisition,''
\newblock in {\em ACM TOG}. ACM, 2008, vol.~27, p.~55.

\bibitem{babacan2012compressive}
S~Derin Babacan, Reto Ansorge, Martin Luessi, Pablo~Ruiz Mataran, Rafael
  Molina, and Aggelos~K Katsaggelos,
\newblock ``Compressive light field sensing,''
\newblock {\em IEEE TIP}, vol. 21, no. 12, pp. 4746--4757, 2012.

\bibitem{inagaki2018learning}
Yasutaka Inagaki, Yuto Kobayashi, Keita Takahashi, Toshiaki Fujii, and Hajime
  Nagahara,
\newblock ``Learning to capture light fields through a coded aperture camera,''
\newblock in {\em ECCV}. Springer, 2018, pp. 431--448.

\bibitem{levin2010linear}
Anat Levin and Fredo Durand,
\newblock ``Linear view synthesis using a dimensionality gap light field
  prior,''
\newblock in {\em CVPR}. IEEE, 2010, pp. 1831--1838.

\bibitem{mousnier2015partial}
Antoine Mousnier, Elif Vural, and Christine Guillemot,
\newblock ``Partial light field tomographic reconstruction from a fixed-camera
  focal stack,''
\newblock {\em arXiv preprint arXiv:1503.01903}, 2015.

\bibitem{srinivasan2017learning}
Pratul~P Srinivasan, Tongzhou Wang, Ashwin Sreelal, Ravi Ramamoorthi, and Ren
  Ng,
\newblock ``Learning to synthesize a 4d rgbd light field from a single image,''
\newblock in {\em ICCV}, 2017, pp. 2243--2251.

\bibitem{levin2007image}
Anat Levin, Rob Fergus, Fr{\'e}do Durand, and William~T Freeman,
\newblock ``Image and depth from a conventional camera with a coded aperture,''
\newblock {\em ACM TOG}, vol. 26, no. 3, pp. 70, 2007.

\bibitem{veeraraghavan2007dappled}
Ashok Veeraraghavan, Ramesh Raskar, Amit Agrawal, Ankit Mohan, and Jack
  Tumblin,
\newblock ``Dappled photography: Mask enhanced cameras for heterodyned light
  fields and coded aperture refocusing,''
\newblock {\em ACM TOG}, vol. 26, no. 3, pp. 69, 2007.

\bibitem{mao2016image}
Xiaojiao Mao, Chunhua Shen, and Yu-Bin Yang,
\newblock ``Image restoration using very deep convolutional encoder-decoder
  networks with symmetric skip connections,''
\newblock in {\em Advances in NIPS}, 2016, pp. 2802--2810.

\bibitem{srinivasan2018aperture}
Pratul~P Srinivasan, Rahul Garg, Neal Wadhwa, Ren Ng, and Jonathan~T Barron,
\newblock ``Aperture supervision for monocular depth estimation,''
\newblock in {\em CVPR}, 2018, pp. 6393--6401.

\bibitem{shin2018epinet}
Changha Shin, Hae-Gon Jeon, Youngjin Yoon, In~So Kweon, and Seon~Joo Kim,
\newblock ``Epinet: A fully-convolutional neural network using epipolar
  geometry for depth from light field images,''
\newblock in {\em CVPR}, 2018, pp. 4748--4757.

\bibitem{nabati2018fast}
O.~Nabati, D.~Mendlovic, and R.~Giryes,
\newblock ``Fast and accurate reconstruction of compressed color light field,''
\newblock in {\em ICCP}, May 2018, pp. 1--11.

\bibitem{lippmann}
Gabriel Lippman,
\newblock ``La photograhie integrale,''
\newblock in {\em Academic des Sciences}. ACM, 1908, pp. 31--42.

\bibitem{levoy1996light}
Marc Levoy and Pat Hanrahan,
\newblock ``Light field rendering,''
\newblock in {\em Proc. of the 23rd annual conference on Computer graphics and
  interactive techniques}. ACM, 1996, pp. 31--42.

\bibitem{ng2005light}
Ren Ng, Marc Levoy, Mathieu Br{\'e}dif, Gene Duval, Mark Horowitz, and Pat
  Hanrahan,
\newblock ``Light field photography with a hand-held plenoptic camera,''
\newblock {\em CSTR}, vol. 2, no. 11, pp. 1--11, 2005.

\bibitem{lytro}
Lytro,
\newblock ``https://www.lytro.com/imaging,'' 2017.

\bibitem{raytrix}
Raytrix,
\newblock ``https://raytrix.de/,'' 2019.

\bibitem{zhang2015light}
Zhoutong Zhang, Yebin Liu, and Qionghai Dai,
\newblock ``Light field from micro-baseline image pair,''
\newblock in {\em CVPR}, 2015, pp. 3800--3809.

\bibitem{didyk2013joint}
Piotr Didyk, Pitchaya Sitthi-Amorn, William Freeman, Fr{\'e}do Durand, and
  Wojciech Matusik,
\newblock ``Joint view expansion and filtering for automultiscopic 3d
  displays,''
\newblock {\em ACM TOG}, vol. 32, no. 6, pp. 221, 2013.

\bibitem{yoon2015learning}
Youngjin Yoon, Hae-Gon Jeon, Donggeun Yoo, Joon-Young Lee, and In~So~Kweon,
\newblock ``Learning a deep convolutional network for light-field image
  super-resolution,''
\newblock in {\em ICCVW}, 2015, pp. 24--32.

\bibitem{gul2018spatial}
M~Shahzeb~Khan Gul and Bahadir~K Gunturk,
\newblock ``Spatial and angular resolution enhancement of light fields using
  convolutional neural networks,''
\newblock {\em IEEE TIP}, vol. 27, no. 5, pp. 2146--2159, 2018.

\bibitem{wu2017light}
Gaochang Wu, Mandan Zhao, Liangyong Wang, Qionghai Dai, Tianyou Chai, and Yebin
  Liu,
\newblock ``Light field reconstruction using deep convolutional network on
  epi,''
\newblock in {\em CVPR}, 2017, vol. 2017, p.~2.

\bibitem{hirsch2014switchable}
Matthew Hirsch, Sriram Sivaramakrishnan, Suren Jayasuriya, Albert Wang, Alyosha
  Molnar, Ramesh Raskar, and Gordon Wetzstein,
\newblock ``A switchable light field camera architecture with angle sensitive
  pixels and dictionary-based sparse coding,''
\newblock in {\em 2014 IEEE International Conference on Computational
  Photography (ICCP)}. IEEE, 2014, pp. 1--10.

\bibitem{flynn2016deepstereo}
John Flynn, Ivan Neulander, James Philbin, and Noah Snavely,
\newblock ``Deepstereo: Learning to predict new views from the world's
  imagery,''
\newblock in {\em CVPR}, 2016, pp. 5515--5524.

\bibitem{godard2017unsupervised}
Cl{\'e}ment Godard, Oisin Mac~Aodha, and Gabriel~J Brostow,
\newblock ``Unsupervised monocular depth estimation with left-right
  consistency,''
\newblock in {\em CVPR}, 2017, vol.~2, p.~7.

\bibitem{garg2016unsupervised}
Ravi Garg, Gustavo Carneiro, and Ian Reid,
\newblock ``Unsupervised cnn for single view depth estimation: Geometry to the
  rescue,''
\newblock in {\em ECCV}. Springer, 2016, pp. 740--756.

\bibitem{masia2012perceptually}
Belen Masia, Lara Presa, Adrian Corrales, and Diego Gutierrez,
\newblock ``Perceptually optimized coded apertures for defocus deblurring,''
\newblock in {\em Computer Graphics Forum}. Wiley Online Library, 2012,
  vol.~31, pp. 1867--1879.

\bibitem{pentland1987new}
Alex~Paul Pentland,
\newblock ``A new sense for depth of field,''
\newblock {\em Trans. on PAMI}, , no. 4, pp. 523--531, 1987.

\bibitem{rajagopalan1999mrf}
Ambasamudram~N Rajagopalan and Subhasis Chaudhuri,
\newblock ``An mrf model-based approach to simultaneous recovery of depth and
  restoration from defocused images,''
\newblock {\em Trans. on PAMI}, vol. 21, no. 7, pp. 577--589, 1999.

\bibitem{hasinoff2009multiple}
Samuel~W Hasinoff and Kiriakos~N Kutulakos,
\newblock ``Multiple-aperture photography for high dynamic range and
  post-capture refocusing,''
\newblock {\em IEEE Trans. on PAMI}, vol. 1, no. 1, pp. 1, 2009.

\bibitem{vadathya2018learning}
Anil~Kumar Vadathya, Saikiran Cholleti, Gautham Ramajayam, Vijayalakshmi
  Kanchana, and Kaushik Mitra,
\newblock ``Learning light field reconstruction from a single coded image,''
\newblock {\em arXiv preprint arXiv:1801.06710}, 2018.

\bibitem{xu2012high}
Zhimin Xu and Edmund~Y Lam,
\newblock ``A high-resolution lightfield camera with dual-mask design,''
\newblock in {\em Image Reconstruction from Incomplete Data VII}. International
  Society for Optics and Photonics, 2012, vol. 8500, p. 85000U.

\bibitem{yu2015multi}
Fisher Yu and Vladlen Koltun,
\newblock ``Multi-scale context aggregation by dilated convolutions,''
\newblock in {\em 4th ICLR}, 2016.

\bibitem{kingma2014adam}
Diederik Kingma and Jimmy Ba,
\newblock ``Adam: A method for stochastic optimization,''
\newblock {\em arXiv preprint arXiv:1412.6980}, 2014.

\bibitem{stanfordlytroLF2016}
Lightfields.stanford.edu,
\newblock ``http://lightfields.stanford.edu/lf2016.html,'' 2019.

\end{thebibliography}

%




\end{document}